\documentclass{article}

    \usepackage[preprint]{neurips_2024}

\usepackage[utf8]{inputenc} %
\usepackage[T1]{fontenc}    %
\usepackage{hyperref}       %
\usepackage{url}            %
\usepackage{booktabs}       %
\usepackage{amsfonts}       %
\usepackage{nicefrac}       %
\usepackage{microtype}      %
\usepackage[dvipsnames]{xcolor}         %
\usepackage{multirow}
\usepackage{wrapfig}
\usepackage{listings}

\usepackage{graphicx}
\usepackage{enumitem}
\setlist{nosep}
\usepackage{soul}
\usepackage{tabularx}
\usepackage{subcaption}

\newcommand{\jasonhidden}[1]{}

\title{AnnotatedTables: A Large Tabular Dataset with Language Model Annotations}

\author{%
    Yaojie Hu \\
    Department of Computer Science \\
    Iowa State University \\ 
    \texttt{jhu@iastate.edu}
    \And
    Ilias Fountalis \\
    RelationalAI \\
    \texttt{ilias.fountalis@relational.ai}
    \And
    Jin Tian \\
    Department of Computer Science \\
    Iowa State University \\ 
    \texttt{jtian@iastate.edu}
    \And
    Nikolaos Vasiloglou \\
    RelationalAI \\
    \texttt{nik.vasiloglou@relational.ai}
}

\begin{document}

\maketitle

\begin{abstract}
Tabular data is ubiquitous in real-world applications and abundant on the web, yet its annotation has traditionally required human labor, posing a significant scalability bottleneck for tabular machine learning.
In this paper, we introduce methods that leverage Large Language Models (LLMs) to understand and annotate tabular data.
Our methodology can successfully annotate a large amount of tabular data and can be flexibly steered to generate various types of annotations based on specific research objectives, as we demonstrate with SQL annotation and input-target column annotation as examples.
As a result, we release AnnotatedTables, a collection of 32,119 databases with LLM-generated annotations. The dataset includes 405,616 valid SQL programs, making it the largest SQL dataset with associated tabular data that supports query execution.
To further demonstrate the value of our methodology and dataset, we perform two follow-up research studies.
1) We investigate whether LLMs can translate SQL programs to Rel programs, a database language previously unknown to LLMs, while obtaining the same execution results. Using our Incremental Prompt Engineering methods based on execution feedback, we show that LLMs can produce adequate translations with few-shot learning.
2) We evaluate the performance of TabPFN, a recent neural tabular classifier trained on Bayesian priors, on 2,720 tables with input-target columns identified and annotated by LLMs. 
On average, TabPFN performs on par with the baseline AutoML method, though the relative performance can vary significantly from one data table to another, making both models viable for practical applications depending on the situation. 
Our findings underscore the potential of LLMs in automating the annotation of large volumes of diverse tabular data.

\end{abstract}

\section{Introduction}

Tabular data is one of the most common data types with pervasive real-world applications such as finance, healthcare management, and marketing analytics \citep{shwartz2022tabular, borisov2022deep, chui2018notes}. The tabular format can organize structured data as relational databases.
Machine learning models are able to understand and generate SQL queries against the relational databases \citep{zhong2017seq2sql, yan2020sql}, and large language models (LLMs) have emerged as a prominent method in this field \citep{liu2023comprehensive, li2024can}. Recent studies show that LLMs can achieve state-of-the-art SQL generation with carefully engineered prompts \citep{gao2023text}, owing thanks to the massive amount of open-source SQL code in their training set \citep{li2023starcoder}. 
Although valuable for training LLMs, SQL code alone is not enough for other tabular machine learning problems. For example, to evaluate the text-to-SQL ability where the model generates SQL code based on a description, both tabular data and SQL code would be required to calculate the execution accuracy \citep{zhong2017seq2sql, xu2017sqlnet, chang2023how}. 
While either SQL code or tabular data can be abundant, obtaining both at the same time can be far more difficult. Not limited to SQL code, annotations for tabular datasets can be generally difficult to collect, which we call \textit{tabular dataset annotation problem} in this paper. We use executable SQL code as a lens to explore this tabular dataset annotation problem in detail and propose our machine learning solutions.

WikiSQL \citep{zhong2017seq2sql} and Spider \citep{yu2018spider} are two popular datasets with both SQL code and tabular data suitable for execution. However, their sizes are limited, mainly because their annotation methodologies do not scale up to make full use of either the tabular data or the SQL code on the web.
WikiSQL gathers its tabular data from Wikipedia pages, which is a limited source and does not reflect the complications of real-world tabular data. Also, its SQL programs are annotated manually by a crowd-sourcing service called Amazon Mechanical Turk. 
Every crowd-sourced SQL program in WikiSQL only queries one table at a time, without table-join operations. 
Spider, on the other hand, does include relatively more complex queries on multiple tables in a database, but it also relies on human annotations from 11 college students to collectively write the SQL programs. Human annotation is common in all tabular datasets with executable SQL code prior to our work (Table \ref{tab:count}), an expensive methodology to reproduce and expand upon \citep{hsueh2009data}.
However, recent work shows that LLMs are able to synthesize SQL code robustly given instructions \citep{liu2023comprehensive}, which motivates us to explore the potential to use LLMs to synthesize the SQL code for the abundant, unannotated tabular data. 
Our work focuses on SQL code, but  SQL code is not the only type of annotation that tabular machine learning systems may need. Can LLMs create other types of annotations and generally replace expensive human annotations for tabular datasets?
Inspired by the multi-tasking \citep{srivastava2022beyond} and emergent abilities \citep{wei2022emergent, schaeffer2024emergent} of LLMs, we see both value and potential in tackling the tabular dataset annotation problem.

\begin{wraptable}[14]{r}{0.45\textwidth}
\vskip -13pt
\caption{The sizes of the executable SQL datasets in the literature compared to ours.} %
    \centering
    \vskip -6pt
        \resizebox{\linewidth}{!}{%
    \begin{tabular}{lrr}
\toprule
         Dataset & SQL & DB  \\
\midrule
         WikiSQL \citep{zhong2017seq2sql} & 80654 & 26251\\
         Spider \citep{yu2018spider} & 9693 & 200 \\
         Cspider \citep{min-etal-2019-pilot} & 9691 & 166 \\
         IMDB \citep{yaghmazadeh2017sqlizer} & 111 & 1\\
         Yelp \citep{yaghmazadeh2017sqlizer} & 68 & 1 \\
         SparC \citep{yu-etal-2019-sparc}  & 15598 & 200\\
         TableQA \citep{sun2020tableqa} & 49974 & 5291 \\
         DuSQL \citep{wang-etal-2020-dusql} & 23797 & 200 \\
         KaggleDBQA \citep{lee-etal-2021-kaggledbqa} & 272 & 8 \\
         Chase \citep{guo2021chase} & 17940 & 280 \\
         BIRD \citep{li2024can} & 12751 & 95 \\ 
     \midrule
         AnnotatedTables (Ours) & 405616 & 32119 \\
\bottomrule
    \end{tabular}
    }
    \label{tab:count}
\end{wraptable}

In this paper, we introduce AnnotatedTables, a large scale tabular dataset with large language model annotations. 
To overcome the bottleneck of labor-intensive human annotations, we use LLMs to understand tabular data and generate synthetic annotations. 
Our dataset construction starts with diverse cross-domain tabular data 
encountered in real-world data science practices, while maintaining general applicability to other tabular data sources.
With careful prompt design, we instruct LLMs to synthesize SQL code with zero-shot learning, which, despite its large scale, proves to be a quick and cost-effective method to create a rich dataset without expensive human labor.
To ensure the quality of LLM-generated SQL code, we verify them with execution on the tabular data, and we report the validity of generated SQL queries for analysis.
To date, AnnotatedTables is the largest collection of SQL code that can be executed on the associated tabular data (Table \ref{tab:count}). 
SQL programs are not the only annotations LLMs can produce. Some columns in a table may serve as model input to predict others, and such input-target columns are usually identified by people manually. 
The LLM annotation methodology can be extended to annotate input-target columns for our tables, making them suitable %
for tabular classification. 
Our work shows that LLMs can simplify and automate the process of tabular dataset construction.

To validate the soundness and value of our dataset, we conduct two follow-up studies. 1) We study LLM's ability to learn a new programming language in-context. 2) We scale up the evaluation of a recent tabular classification model and test its performance on diverse real-world data. Both studies demonstrate how LLM annotation can flexibly adapt our dataset to target specific research problems. Our code and data will be released publicly.

The contributions of our paper are summarized as follows.
\begin{itemize}[itemsep=0.2ex,partopsep=0ex,parsep=.2ex, leftmargin=2ex,topsep=-1ex]
    \item We introduce in-context learning methods of LLMs to understand and annotate a large amount of real-world tabular data, %
    addressing the scaling bottleneck of traditional human annotation.
    \item We use a LLM to annotate public data tables and construct a cross-domain dataset named AnnotatedTables, currently the largest SQL dataset with tabular data that supports query execution.
    \item We extensively assess the quality of the LLM-generated annotations throughout the study. We demonstrate the research value of AnnotatedTables dataset through two follow-up studies.
    \item We introduce Incremental Prompt Engineering to teach LLMs to translate SQL programs to Rel programs with adequate accuracy, where Rel is a new language not previously learned by LLMs. 
    \item We use the input-target columns annotated by LLMs to study TabPFN's performance on %
    a wide variety of tabular classification problems. 
    We find that TabPFN performs overall on par with the baseline method, with each model potentially performing better depending on the tables.
\end{itemize}

\section{Related Work}

\paragraph{Database management systems and deep learning}
For text-to-SQL, language models can be trained or fine-tuned to synthesize SQL queries given natural language descriptions 
\citep{yavuz2018takes, hwang2019comprehensive, yu2018typesql}. 
For cardinality estimation, neural networks can predict the number of rows returned by an intermediate operation for query optimization \citep{liu2015cardinality, negi2020cost,woltmann2019cardinality, yang2019deep, yang2020neurocard},
and end-to-end neural query optimizers can be built %
\citep{fischer2019dl2, gupta2020deeprecsys, marcus2019neo, marcus2021bao}.
Other deep learning for database problems
include query answering \citep{hilprecht2019deepdb},
anomaly detection \citep{pang2021deep},
ontology matching \citep{bento2020ontology},
relational data modeling \citep{fan2020relational, liu2022learning}, and database indexing \citep{kraska2018case}.
All these problems could potentially benefit from a large-scale multi-domain dataset like AnnotatedTables. LLM annotation can also be applied to these problems to acquire a large amount of tabular training data. Moreover, our dataset supports query execution, which is required for text-to-SQL and cardinality estimation to collect ground truth.

\paragraph{SQL datasets with tabular data.}
SQL annotation is a focus of our paper. AnnotatedTables is the largest SQL dataset with associated tabular data that supports query execution. 
Currently, the most commonly used SQL datasets with tabular data are listed in Table~\ref{tab:count}, surveyed in \citep{deng-etal-2022-recent, qin2022survey, li2024can}. All of them are annotated by humans, and our work is the first tabular dataset annotated by LLMs. Our LLM methods can cheaply generate a large amount of annotations that match the volume of tabular data available on the web.
Existing SQL datasets with tabular data often include English descriptions for text-to-SQL evaluation. While we collect these descriptions, we do not assess or ensure their quality. Unlike SQL programs, which can be validated through execution \citep{yu-etal-2019-sparc}, English descriptions cannot be executed.

\paragraph{Large Language Models for data annotation.}

LLMs have been used for data annotation on text domains \citep{mao-etal-2024-gpteval-survey} and can outperform human annoators \citep{gilardi2023chatgpt}. LLMs can detect implicit hate speech and provide natural language explanations as annotations \citep{huang2023chatgpt}. 
LLMs are also used for natural language generation evaluation, showing a high alignment with human evaluation \citep{liu-etal-2023-g}.
Texts can be directly understood and annotated by LLMs, while tabular data requires additional annotation methods, which we introduce.

\section{AnnotatedTables}

\subsection{Dataset Construction}

\label{sec:construct}

\paragraph{Collection of public data tables}
The construction of AnnotatedTables starts by collecting public tabular data.
Our goal is to tap into the large quantity of un-annotated tabular data that exist on the web, with a preference for curated tables encountered in real-world data science applications.
Kaggle, a data science platform, meets our requirements \citep{megan_risdal_timo_bozsolik_2022, quaranta2021kgtorrent,wang2021makes}.
Kaggle hosts data tables that are crowd sourced by a community of data enthusiasts and professionals. Users can share and analyze datasets on a wide variety of real-world topics, including football players, U.S airline flights, renewable energy output, weather forecasting, and e-commerce product reviews.
Users may rate, improve, and analyze the datasets on Kaggle, contributing to the curation of useable and reliable data.
Although individual Kaggle datasets have been studied in prior works \citep{bojer2021kaggle, taieb2014gradient, yang2018deep, tolkachev2020deep, yang2020computational}, our research aggregates and examines the data tables en masse, uncovering an previously overlooked goldmine of diverse and high-quality data tables.

We use Kaggle's public API to search for datasets
with sizes ranging from 10 megabytes to 1 gigabyte for faster data collection and processing.
We also require every dataset's ``usability rating'' to be higher than 0.1. The ``usability rating'' is a quality score given by Kaggle users and helps us filter out low-quality datasets and datasets with no usages.
We collect 70,000 Kaggle datasets before stopping for further processing.
Our LLM annotation methodology is flexible and automated. The rest of our methods can be applied to other data sources and is not limited to Kaggle datasets alone.

\paragraph{Describing databases to LLMs with schema and example rows}
In order for LLMs to understand and annotate tabular data, we need to design a textual description for each database as a part of the prompt.
Using entire tables in the prompt could exceed the maximum input length of LLMs, and we  use a much shorter description instead that contains a basic schema and example rows.

\begin{figure}[t]
    \centering
    \includegraphics[width=\linewidth]{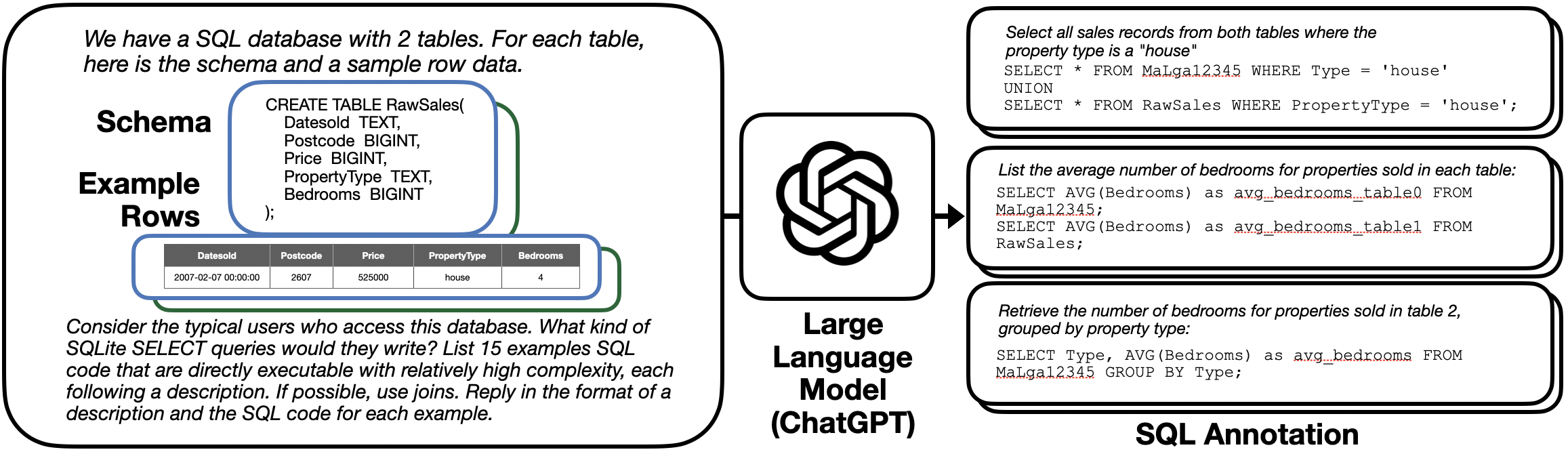}
    \vskip 0pt
    \caption{An illustration of the SQL code annotation process with a large language model.
    }
    \vskip -15pt
    \label{fig:sqlprompt}
\end{figure}

Table schema is a summary description of its structure, commonly used to define the table during its creation \citep{uschold2015ontology}.
For our annotation purposes, we use a basic schema that includes the table name, the table's column names, and the data type of each column (Figure \ref{fig:sqlprompt}).
We find tabular data stored as CSV files in the Kaggle datasets and automate schema extraction.
As CSV files do not include table names in their content, we opt to use the file names as table names.
Column names are included in the CSV files as the header row.
Upon inspection, we find that the column names and file names are typically well-labeled and descriptive enough %
for LLMs to infer the information in the table.
Lastly, %
we use an existing data analysis software (pandas) to load the tables and infer the column data types.
Our schema extraction method is practical and effective.

A Kaggle dataset may contain multiple tables and therefore multiple table schemas. The set of %
table schemas describe the database structure: how many tables there are, %
what the columns and types are, and whether a column appears in multiple tables that could potentially be a \texttt{JOIN} key.

Example rows contain content from the table, and they are necessary for LLMs to write many common SQL queries, such as \texttt{WHERE} statements that filter the table based on a condition. For a concrete example, in \texttt{SELECT * FROM CUSTOMERS WHERE COUNTRY=`MEXICO'}, the table name \texttt{CUSTOMERS} and column name \texttt{COUNTRY} can be found in the database schemas, while  \texttt{`MEXICO'} is a value that can only be found in a table row. Providing row examples in addition to the schema is necessary for LLMs to synthesize \texttt{WHERE} statements in SQL.
For the example row, we use the
first row of each table, converted to text through JSON formatting.
Our schema and example row descriptions are generic and allow almost all tabular data to be annotated by LLMs.

\paragraph{SQL annotations with zero-shot synthesis by LLMs}

Through zero-shot learning, we instruct LLMs to synthesize SQL code that queries the data tables we collected as one type of annotation. Currently, OpenAI's ChatGPT  \citep{ouyang2022training} is one of the best performing LLMs on various benchmarks \citep{hendrycks2021measuring, zellers-etal-2019-hellaswag, sakaguchi2021winogrande}.
For our paper, we use ChatGPT with GPT-3.5 backbone as our annotation LLM, referred to as ``the LLM'' from now on.

To perform zero-shot SQL annotation for each database, we create a prompt that contains the instruction, the database schema, and the example row of every table, illustrated in Figure \ref{fig:sqlprompt}.
The instruction asks the LLM to write SQL queries from ``typical users who access this database''. In addition, the instruction asks for queries ``with relatively high complexity'', which steers the synthesis results to more complicated SQL code and makes the dataset more challenging. Lastly, the instruction asks the language model to create more \texttt{JOIN} queries when multiple tables are present, as \texttt{JOIN} queries are especially valuable data points when studying problems including cardinality estimation and query optimization \citep{yang2020neurocard, marcus2021bao}.
For every Kaggle dataset with tabular data, we prompt the LLM to generate 15
SQL program annotations with natural language annotations.

\paragraph{Execution-based validation}
Some of the LLM-generated SQL programs could be incorrect, and it is important to examine and ensure the quality of the LLM annotations.
As AnnotatedTables has tabular data available for execution, we verify the validity of the SQL programs by executing them \citep{yu-etal-2019-sparc}. If a SQL program can be executed and returns non-empty results, then we classify it as a valid SQL annotation. Executable queries with empty results are marked. In our follow-up study of SQL-to-Rel translation, a pair of SQL and Rel queries with empty results cannot be compared by execution accuracy, and SQL queries with empty results will not be used.
We do not validate the English descriptions. All intermediate data artifacts are released as a part of AnnotatedTables.

\begin{table}[]
\caption{Examples of LLM-generated English and SQL annotations for Kaggle tabular data. LLMs can generate SQL programs that cannot be executed, and we mark the incorrect parts in red.}
    \centering
\resizebox{\textwidth}{!}{%
\begin{tabular}{lp{11cm}p{10cm}p{2cm}}
\toprule
 & English Annotation & SQL Annotation & Execution \\
\midrule
1 & Get the average humidity for each location in the morning and afternoon: & \texttt{SELECT Location, AVG(Humidity9am) AS AverageHumidity9am, AVG(Humidity3pm) AS AverageHumidity3pm
FROM WeatherAus
GROUP BY Location;} & Valid \\
2 & Retrieve rows where the tempo is greater than the average tempo: & \texttt{SELECT * FROM LikedsongsGV WHERE Tempo > (SELECT AVG(Tempo) FROM LikedsongsGV);} & Valid \\
3 & Retrieve records from a table that meet a certain condition: & \texttt{SELECT * FROM \textcolor{red}{table\_name} WHERE \textcolor{red}{condition};} & Error \\
4 & Retrieve rows where column 2 is less than 0.1 and column 5 is greater than 0.2: & \texttt{SELECT * FROM Preds001 WHERE \textcolor{red}{2} < 0.1 AND \textcolor{red}{5} > 0.2;} & Error \\
5 & Get the total number of matches played in a specific season & \texttt{SELECT COUNT(*) FROM IPLMatches WHERE Season = 2019;} & Empty Result \\
\bottomrule
\end{tabular}
}
\vskip -15pt
    \label{tab:sql-examples}
\end{table}

\begin{wraptable}[17]{r}{0.5\textwidth}
\vskip -20pt
    \caption{SQL code synthesized by LLM categorized by the types of SQL components.
    }
    \vskip -5pt
    \centering
    \resizebox{\linewidth}{!}{%
\begin{tabular}{lrrrrr}
\toprule
Component & Total & Valid & \% Valid & \% Error & \% Empty \\
\midrule
Total & 493134 & 405616 & 82.25 & 9.96 & 7.79 \\
\midrule
\texttt{FROM} & 493093 & 405612 & 82.26 & 7.78 & 9.96 \\
\texttt{WHERE} & 276710 & 209393 & 75.67 & 8.07 & 16.26 \\
\texttt{AND} & 88182 & 61814 & 70.10 & 10.81 & 19.09 \\
\texttt{GROUP BY} & 77004 & 66061 & 85.79 & 9.74 & 4.47 \\
\texttt{AVG} & 65797 & 58238 & 88.51 & 8.83 & 2.66 \\
\texttt{COUNT} & 58586 & 53486 & 91.29 & 6.41 & 2.29 \\
\texttt{ORDER BY} & 52677 & 45941 & 87.21 & 8.98 & 3.81 \\
\texttt{LIKE} & 38865 & 28189 & 72.53 & 5.90 & 21.57 \\
\texttt{JOIN} & 30079 & 14879 & 49.47 & 23.00 & 27.53 \\
\texttt{MAX} & 29693 & 26188 & 88.20 & 9.29 & 2.52 \\
\texttt{SUM} & 24801 & 21324 & 85.98 & 10.14 & 3.87 \\
\texttt{LIMIT} & 24241 & 21573 & 88.99 & 8.23 & 2.77 \\
\texttt{NOT} & 19180 & 15111 & 78.79 & 7.22 & 13.99 \\
\texttt{IN} & 16011 & 9656 & 60.31 & 23.76 & 15.93 \\
\texttt{MIN} & 12368 & 10938 & 88.44 & 9.38 & 2.18 \\
\texttt{OR} & 11199 & 8346 & 74.52 & 11.09 & 14.39 \\
\texttt{UNION} & 10408 & 7375 & 70.86 & 24.95 & 4.19 \\
\texttt{HAVING} & 2507 & 1427 & 56.92 & 15.96 & 27.12 \\
\texttt{INTERSECT} & 184 & 101 & 54.89 & 10.33 & 34.78 \\
\texttt{EXCEPT} & 37 & 26 & 70.27 & 13.51 & 16.22 \\
\bottomrule
\end{tabular}
}
    \label{tab:sql}
\end{wraptable}

\subsection{Features, Statistics, and Examples of AnnotatedTables}
\paragraph{Human-like annotation} We find that the LLM can write SQL programs with human-like intent and usage. Examples of the LLM-annotated SQL programs are provided in Table~\ref{tab:sql-examples}. Instructed by our prompt to play the role of a typical user of a database, the LLM writes meaningful queries to look for useful data analytics.

\paragraph{Correctness and complexity} %
Given a high percentage (82.25\%) of valid SQL programs, the annotation quality is good. The LLM can write complex programs with a variety of SQL components (Table \ref{tab:sql}).
The percentages of valid queries in all SQL component categories except \texttt{JOIN} exceed 50\%.
49.47\% of \texttt{JOIN} queries are valid, possibly reflecting a higher level of data understanding and programming skills required to combine multiple tables.

For limitations, we note that some simple SQL statements appear often, in the form of \texttt{SELECT * FROM table}, reflected by a high percentage (48.4\% among valid SQL programs) of \texttt{FROM} statements without other SQL components. Although valid, they may not be as valuable as the more complicated SQL programs for downstream learning and analysis. In the prompt, we steer the LLM to generate SQL programs ``with relative high complexity'' and see success to some extent (Appendix \ref{appx:sql_steering_complex}). However, the dual objectives to generate complex queries and correct queries may inherently conflict with each other, and LLMs could be inclined to generate simple ones to prioritize correctness over complexity. %

\paragraph{Invalid SQL annotations} As shown in Table \ref{tab:sql-examples}, some LLM-annotated SQL programs may be invalid for different reasons. At times, the LLM hallucinates non-existing table names such as ``\texttt{table\_name}'' (Example 3) or leaves a place holder text ``\texttt{condition}'' instead of the exact selection condition (Example 3). In some CSV files, the column names may be numerical, which are invalid column names in a SQL statement, and the LLM is unable to annotate (Example 4). Lastly, as execution result needs to be non-empty so that execution accuracy can be computed for downstream tasks, some syntactically correct SQL programs that return empty results are labeled (Example 5). %

\section{AnnotatedTables for In-Context Translation from SQL to Rel}
\label{sec:rel}

We use AnnotatedTables to study large language models' ability to learn to code in Rel, a new database language \citep{RelationalAI_2021}. 
For new programming languages, there is not much open-source code 
available to train machine learning models with gradient-descent methods,
and, with few training examples, few-shot in-context learning (ICL) becomes a natural candidate instead
\citep{min2022rethinking}. 
ICL has been studied, but ICL of a new programming language is a novel research problem. New programming languages are rare, and the lens of this study is unique and timely.

All SQL programs in AnnotatedTables are executable, which is necessary in order to evaluate the correctness of the translations with execution accuracy.
AnnotatedTables contains many SQL programs and data tables, which can measure the general performance of our SQL-to-Rel translation method.
SQL programs in AnnotatedTables emulate the usage and intent of typical database users, and, by translating these queries, we indirectly synthesize a large dataset of non-trivial Rel programs that is valuable for downstream analysis and machine learning of the new programming language. 

Machine learning of a new programming language may seem challenging with limited training data, but as our results show, our translation accuracy increases steadily as more in-context examples are included, and with 45 translation pairs in the prompt, the translation accuracy can reach around 40\%. %

\subsection{Task Formulation of SQL-to-Rel Translation}

Given a SQL program $S$ that can be executed on table $T$, the task is to build a model $\mathcal M$ that
learns to translate $S$ into a Rel program $R$, such that $S$ and $R$ return the same query results when executed on table $T$. In our case, the model $\mathcal M$ is a LLM parameterized by an instruction prompt that contains few-shot examples. Execution accuracy is the percentage of queries $S, R$ that return the same non-empty results.
Execution accuracy is a common metric for SQL generation \citep{zhong2017seq2sql, xu2017sqlnet, chang2023how}. 
Program equivalence is generally undecidable \citep{rice1953classes, strichman2018program}, and it is not possible to directly compare SQL and Rel programs.

\subsection{Extending AnnotatedTables with Few-shot SQL-to-Rel Translation Pairs}

While LLMs have already learned to program in SQL from comprehensive pre-training \citep{brown2020language,openai2023gpt4}, they %
do not know how to write programs in Rel. To enable ICL, one idea is to use the Rel language specification and documentation as the prompt. %
However, language models such as ChatGPT have limited context span up to 16,000 tokens (around 12,000 English words) \citep{tay2020long, beltagy2020longformer}, and it is currently infeasible for the prompt to contain the complete Rel specification and documentation (estimated 500,000 words). Instead, we use SQL-to-Rel translation examples to teach the LLM to code in Rel, which is simple and effective. Rel is a declarative language similar to Datalog and Prolog. LLMs may transfer their background knowledge in related languages to learn Rel. ICL through translation examples cannot comprehensively cover all the edge cases of the Rel language, but, as our results show, around 40\% of our SQL code can be successfully translated.%

\begin{table}[t]
    \caption{Examples of the SQL-to-Rel translations produced, categorized by ``correct translations'', ``execution errors'', and ``different results''. 
    }
    \centering
\resizebox{\textwidth}{!}{%
\begin{tabular}{lp{8cm}p{13.5cm}p{2cm}}
\toprule
& SQL & Rel Translated & Ex. Acc. \\
\midrule
1 & \texttt{SELECT Departments, MAX(AverageMontlyHours) AS MaxMonthlyHours FROM Attrition GROUP BY Departments;} & \texttt{def output[departments] = max[row\_id, avg\_monthly\_hours: Attrition:AverageMontlyHours(row\_id, avg\_monthly\_hours) and Attrition:Departments[row\_id, departments]]} & Correct \\
\midrule
2 & \texttt{SELECT (SELECT COUNT(*) FROM Oof) + (SELECT COUNT(*) FROM Submission) as TotalCount;} & \texttt{def count\_oof\_rows = count[row\_id: Oof[\_, row\_id, \_]]\newline def count\_submission\_rows = count[row\_id: Submission[\_, row\_id, \_]]\newline def output = count\_oof\_rows + count\_submission\_rows} & Correct \\
\midrule
3 & \texttt{SELECT COUNT(*) FROM DgemFormatValidation;} & \texttt{def count\_rows = count[row\_id: DgemFormatValidation[\_, row\_id, \_]]\newline def output = count\_rows} & Correct \\
\midrule
4 & \texttt{SELECT Education, AVG(Age) AS avg\_age FROM MLPdata GROUP BY Education;} & \texttt{def output[education] = average[row\_id, age: MLPdata:Education[row\_id, education] and MLPdata:Age[row\_id, age]] groupby education} & Error \\
\midrule
5 & \texttt{SELECT COUNT(*) FROM Dict;} & \texttt{def output = count[row, \_ : Dict[\_, row, \_]]} & Error \\
\midrule
6 & \texttt{SELECT * FROM V9d3 ORDER BY Labels DESC;} & \texttt{def output[col, row\_id, val] = V9d3[col, row\_id, val] orderby V9d3:Labels[row\_id] desc} & Error \\
\midrule
7 & \texttt{SELECT MAX(LENGTH(Description)) FROM Cryptosentiment;} & \texttt{def output = max[row\_id, length: Cryptosentiment:Description[row\_id, length]]} & Diff. Results \\
\midrule
8 & \texttt{SELECT * FROM Comments20200605 WHERE ReplyCount >= 5;} & \texttt{def output[col, row\_id, val] = Comments20200605[col, row\_id, val] and Comments20200605:ReplyCount[row\_id] >= 5} & Diff. Results \\
\midrule
9 & \texttt{SELECT AVG(MonthlyRevenue) FROM Cell2celltrain;} & \texttt{def output = average[idx, monthly\_revenue: Cell2celltrain:MonthlyRevenue [idx, monthly\_revenue]]} & Diff. Results \\
\bottomrule
\end{tabular}
}
\vskip -20pt
    \label{tab:rel_examples}
\end{table}

\paragraph{Incremental Prompt Engineering of translation examples} To develop  SQL-to-Rel translation examples to teach LLMs to code in Rel, we introduce a technique called \textit{Incremental Prompt Engineering} (IPE). 
For step $i=1,2,..., k$, we start with some manually annotated SQL-to-Rel translation examples $E_i$ and instruct the language model to translate some unseen SQL programs $S_i$ to Rel programs $R_i$. 
We then calculate the execution accuracy by comparing the query results of $S_i$ and $R_i$ at every step $i$.
The execution accuracy is used as a feedback, and for the SQL code in $S_i$ that the LLM fails to translate, we manually annotate the ones that represent 
common failure patterns to add to $E_i$ as the few-shot examples $E_{i+1}$ at the next step. 
In practice, we start with 5 translation examples and add multiple manual annotations at a time until the execution accuracy converges with around 45 translation examples (Figure \ref{fig:incremental}). 
All examples used in the prompt are in Appendix \ref{appx:sql_rel_examples}.

\subsection{Results}

\begin{wraptable}[16]{r}{0.55\textwidth}
\vskip -40pt
    \centering
    \caption{The LLM's few-shot SQL-to-Rel translation execution results, by the types of SQL components. 
    }
    \vskip -5pt
    \resizebox{\linewidth}{!}{%
\begin{tabular}{lrrrrrr}
\toprule
SQL Comp. & \# Ex. & Translated & Correct & \% Correct & \% Error & \% Diff. \\
\midrule
Total & 45 & 285140 & 116327 & 40.80 & 53.60 & 5.60 \\
\midrule
\texttt{FROM} & 45 & 285136 & 116327 & 40.80 & 53.60 & 5.60 \\
\texttt{WHERE} & 28 & 141893 & 49576 & 34.94 & 59.02 & 6.04 \\
\texttt{GROUP BY} & 9 & 40254 & 7737 & 19.22 & 78.36 & 2.42 \\
\texttt{AND} & 7 & 37963 & 13086 & 34.47 & 59.72 & 5.81 \\
\texttt{AVG} & 5 & 37318 & 12601 & 33.77 & 60.08 & 6.16 \\
\texttt{COUNT} & 6 & 36769 & 17542 & 47.71 & 48.90 & 3.39 \\
\texttt{ORDER BY} & 2 & 28618 & 434 & 1.52 & 97.74 & 0.75 \\
\texttt{LIKE} & 1 & 18399 & 2037 & 11.07 & 82.67 & 6.26 \\
\texttt{MAX} & 2 & 17984 & 7698 & 42.80 & 52.46 & 4.74 \\
\texttt{SUM} & 5 & 14107 & 4077 & 28.90 & 69.69 & 1.41 \\
\texttt{LIMIT} & 0 & 13696 & 73 & 0.53 & 98.53 & 0.94 \\
\texttt{NOT} & 0 & 8754 & 326 & 3.72 & 93.24 & 3.04 \\
\texttt{JOIN} & 8 & 8358 & 526 & 6.29 & 88.32 & 5.38 \\
\texttt{MIN} & 0 & 7344 & 3043 & 41.44 & 55.51 & 3.05 \\
\texttt{IN} & 2 & 5365 & 1045 & 19.48 & 76.07 & 4.45 \\
\texttt{UNION} & 0 & 4530 & 263 & 5.81 & 88.92 & 5.28 \\
\texttt{OR} & 1 & 4071 & 1292 & 31.74 & 60.21 & 8.06 \\
\texttt{HAVING} & 0 & 677 & 1 & 0.15 & 99.26 & 0.59 \\
\texttt{INTERSECT} & 0 & 56 & 2 & 3.57 & 92.86 & 3.57 \\
\texttt{EXCEPT} & 0 & 11 & 1 & 9.09 & 81.82 & 9.09 \\
\bottomrule
\end{tabular}
}
    \label{tab:rel}
\end{wraptable}

\paragraph{The LLM can learn a new programming language adequately from dozens of examples} Table \ref{tab:rel_examples} shows examples of SQL programs and their Rel translations produced by the LLM. As seen in the examples, the two languages differ greatly, and it is not an easy task to translate them, even for experienced programmers. For example, SQL aggregators such as \texttt{MAX} have their counterpart \texttt{max} function in Rel, but the syntax differs greatly (Example 1). Nevertheless, the LLM is able to perform adequate translations by learning a new programming language from just 45 examples in-context. Aggregators including \texttt{MAX}, \texttt{MIN}, and \texttt{COUNT} have higher-than-average translation accuracy. This is an impressive result for a challenging problem. Notably, no manual examples illustrate how to translate \texttt{MIN} queries, and LLM has inferred its usage from \texttt{MAX} examples.

Execution accuracy measures the translation correctness, where every SQL and Rel pair are executed on the same database and compare their results. Table \ref{tab:rel} shows that the overall execution accuracy is 40.8\%, showing that the LLM can generate adequate translations from SQL to Rel. 
Machine learning of new programming languages is challenging and valuable. Since the best text-to-SQL models have 60\%-80\% accuracy \citep{liu2023comprehensive, li2024can}, 40\% SQL-to-Rel is a promising first step, which could be
sufficient for applications such as coding suggestions and automatic test generation. 

\paragraph{Translation errors and limitations}
53.6\% of the Rel translations have errors before returning results, while 5.6\% of the Rel translations finish execution but have different results (Table \ref{tab:rel}). This suggests that most of the failed translations have syntactical problems and cannot be executed.
Examples of inaccurate translations are in Table \ref{tab:rel_examples}. LLMs may use SQL language constructs such as ``\texttt{groupby}'', ``\texttt{orderby}'', and ``\texttt{desc}'' directly in Rel programs (Example 4, 6), causing syntactically incorrect programs. 
In Example 5, the first underscore ``\_'' caused a syntax error.
In Example 7, the SQL method ``\texttt{LENGTH}'' is not demonstrated in-context, and the LLM used ``length'' as a variable name instead, causing a change in semantics and different execution results.
Examples 8 and 9 show different execution results, but the translations are correct by our inspection. Rel is a language based on the Sixth-Normal-Form (6NF) \citep{date2004detailed, knowles20126nf}, which represents all data as sets of tuples with only one value column, without null values or empty rows. Rel's data representation is not the same as SQL's data representation, and we make conservative conversions when comparing the results to prioritize precision over recall by allowing some false negatives in execution accuracy. Therefore, some correct Rel translations may have different results due to the conversion, though such cases are rare, as the percentage of Rel errors is 10x higher than different results. Discounting this limitation, the SQL-to-Rel translation accuracy may be even higher than that reported in Table \ref{tab:rel}.

\begin{wrapfigure}[11]{r}{0.5\textwidth}
    \centering
    \vskip -10pt
    \includegraphics[width=\linewidth]{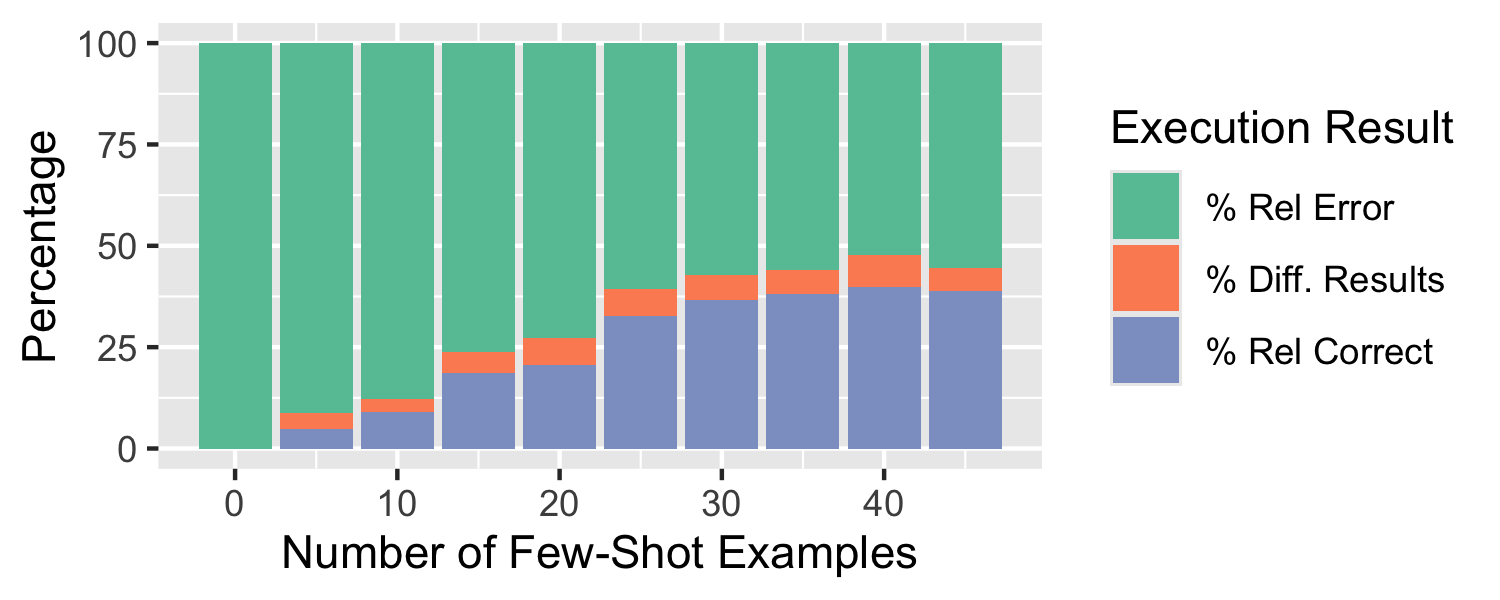}
    \vskip -8pt
    \caption{With Incremental Prompt Engineering, translation accuracy gradually improves and converges as more examples are added.}
    \label{fig:incremental}
\end{wrapfigure}

\paragraph{Performance convergence of Incremental Prompt Engineering}
To study the effectiveness of IPE, we vary the number of few-shot examples incrementally added and measure the translation accuracy on a randomly selected subset of 2,000 SQL programs. We plot the execution accuracy versus the number of translation examples in Figure \ref{fig:incremental}.

Conceptually, Incremental Prompt Engineering adds demonstrations for the failed translations in the previous step and integrates execution feedback into prompt engineering. It is an effective way to improve translation accuracy, as we see a clear increase in the percentage of correctly translated Rel programs as more examples are added. When the number of few-shot examples increases above 30, we see a diminishing return on the accuracy improvements. At 45 examples, the accuracy converges around 40\%. Due to the convergence, more examples may not improve the accuracy beyond 40\%, and additional prompt engineering or LLM techniques may be required.

\section{AnnotatedTables for an Extended Evaluation of TabPFN 
}
\label{sec:pfn}

In this section, we use AnnotatedTables to comprehensively evaluate TabPFN's tabular classification performance on diverse, real-world data tables. 
TabPFN \citep{hollmann2023tabpfn} is a Prior-Data Fitted Network (PFN), a new type of neural networks trained on synthetic data drawn from a Bayesian prior.
Once trained, it takes the entire test set as the input and produces all predictions in a single forward pass. 
TabPFN represents a novel approach to tabular classification, a domain dominated by traditional methods and largely uncontested by deep learning methods \citep{shwartz2022tabular}. 
Inspired by its state-of-the-art results on a small curated benchmark \citep{hollmann2023tabpfn},
we wonder if TabPFN will perform robustly against the %
complications of real-world tabular data.

\paragraph{TabPFN evaluation and ours} TabPFN was evaluated on the OpenML-CC18 benchmark suite \citep{bischl2017openml}. The benchmark has 30 datasets in total, and 18 of them have only numerical features with no missing values, the same as the Bayesian prior that TabPFN is trained on. The authors choose five standard ML methods and two state-of-the-art AutoML methods as baselines to compare with TabPFN. The ML methods include K-nearest-neighbors (KNN), logistic regression, and three gradient boosting methods -- XGBoost \citep{chen2016xgboost}, CatBoost \citep{prokhorenkova2018catboost}, LightGBM \citep{ke2017lightgbm}. The AutoML baseline methods are AutoGluon \citep{erickson2020autogluon, shi2021multimodal} and AutoSkLearn 
\citep{feurer2022auto}. On the 18 numerical datsets, TabPFN outperformed all baselines, and AutoGluon was the best performing baseline (see Table 1 of \citep{hollmann2023tabpfn}), which we will use as the baseline in our study.
In this section, we follow TabPFN authors' experimental settings and use AnnotatedTables to vastly scale up the evaluation, with a focus on assessing its real-world performance.
We evaluate TabPFN on 2,720 tabular classification problems, a 90x increase
over OpenML-CC18.
Moreover, in comparison to the OpenML-CC18 benchmark, AnnotatedTables contains diverse tabular data gathered from a public platform, which would be a more comprehensive evaluation of TabPFN's robustness in real world problems.

\subsection{Task Formulation of Tabular Classification}

For tabular classification, a table $T$ has rows $\{R_1, R_2, ...\}$. A row $R_i$ has one or more column values $\mathbf x_i$ and a categorical column value $y_i$.
A tabular classification model $f$ aims to use $\mathbf x_i$ to predict $y_i$.

Our experimental settings follow the TabPFN authors' settings. 
We train AutoGluon with a time budget of one minute and five minutes per table, and we split the training and test set once per table.
To reduce the total training time as we have many more tables than OpenML-CC18, we do not repeat training on every table with a five-fold training-test set split.
We use AutoGluon as our baseline, because it is the best performing baseline model in the original evaluation \citep{hollmann2023tabpfn}.
Our work does not alter the TabPFN model architecture, and we use the authors' publicly released pre-trained checkpoint. For more model details, we refer readers to the paper \citep{hollmann2023tabpfn}.

\subsection{Extending AnnotatedTables with LLM-Annotated Input-Target Columns}

The Kaggle data tables collected in AnnotatedTables could be used for tabular classification task, if LLMs can identify the potential input and target columns. 
We give the schema description from Sec \ref{sec:construct} to the LLM to identify potential input and target columns in the tables.
TabPFN requires that the input columns are numeric and the target column is categorical for classification. LLM annotation is flexible, and we instruct the LLM to find columns that satisfy this data type requirement (for complete prompt, see Appendix \ref{appx:prompt}).
We use the input-target columns annotated to train TabPFN and AutoGluon, and if the models cannot be trained, the annotations are discarded (Appendix \ref{appx:discard_rate}). 

\subsection{Results}

\begin{table}[t]
\caption{The AUROC (one-versus-one), cross entropy, and time statistics for TabPFN and the AutoGluon baseline on the tabular classification problems in AnnotatedTables.
}
    \centering
        \resizebox{\textwidth}{!}{%
\begin{tabular}{lrrrrrrr}
\toprule
 & & \multicolumn{3}{c}{1 Min (2725 problems)} & \multicolumn{3}{c}{5 Mins (2720 problems)} \\
\cmidrule(lr){3-5} \cmidrule(lr){6-8}
& & TabPFN & AutoGluon & Paired Diff. & TabPFN & AutoGluon & Paired Diff. \\
\midrule
\multirow{3}{*}{AUROC} & Mean & 0.760 $\pm$ 0.173 & 0.723 $\pm$ 0.214 & 0.036 $\pm$ 0.152 & 0.760 $\pm$ 0.174 & 0.755 $\pm$ 0.187 & 0.005 $\pm$ 0.095\\
 & Median & 0.794 & 0.767 & 0.004 & 0.793 & 0.794 & -0.000 \\
\midrule
\multirow{3}{*}{CE} & Mean & 0.753 $\pm$ 0.550 & 0.886 $\pm$ 2.129 & -0.133 $\pm$ 1.944 & 0.753 $\pm$ 0.550 & 0.807 $\pm$ 1.929 & -0.054 $\pm$ 1.726 \\
 & Median & 0.655 & 0.693 & -0.011 & 0.656 & 0.659 & 0.001 \\
\midrule
Time (sec.) & Mean  & 2.002 & 73.654 & -71.652 & 2.342 & 315.038 & -312.696 \\
\bottomrule
\end{tabular}
}
\vskip -13pt
    \label{tab:tabpfn}
\end{table}

\begin{wrapfigure}[16]{r}{0.4\textwidth}
    \centering
    \vskip-42pt
    \includegraphics[width=\linewidth]{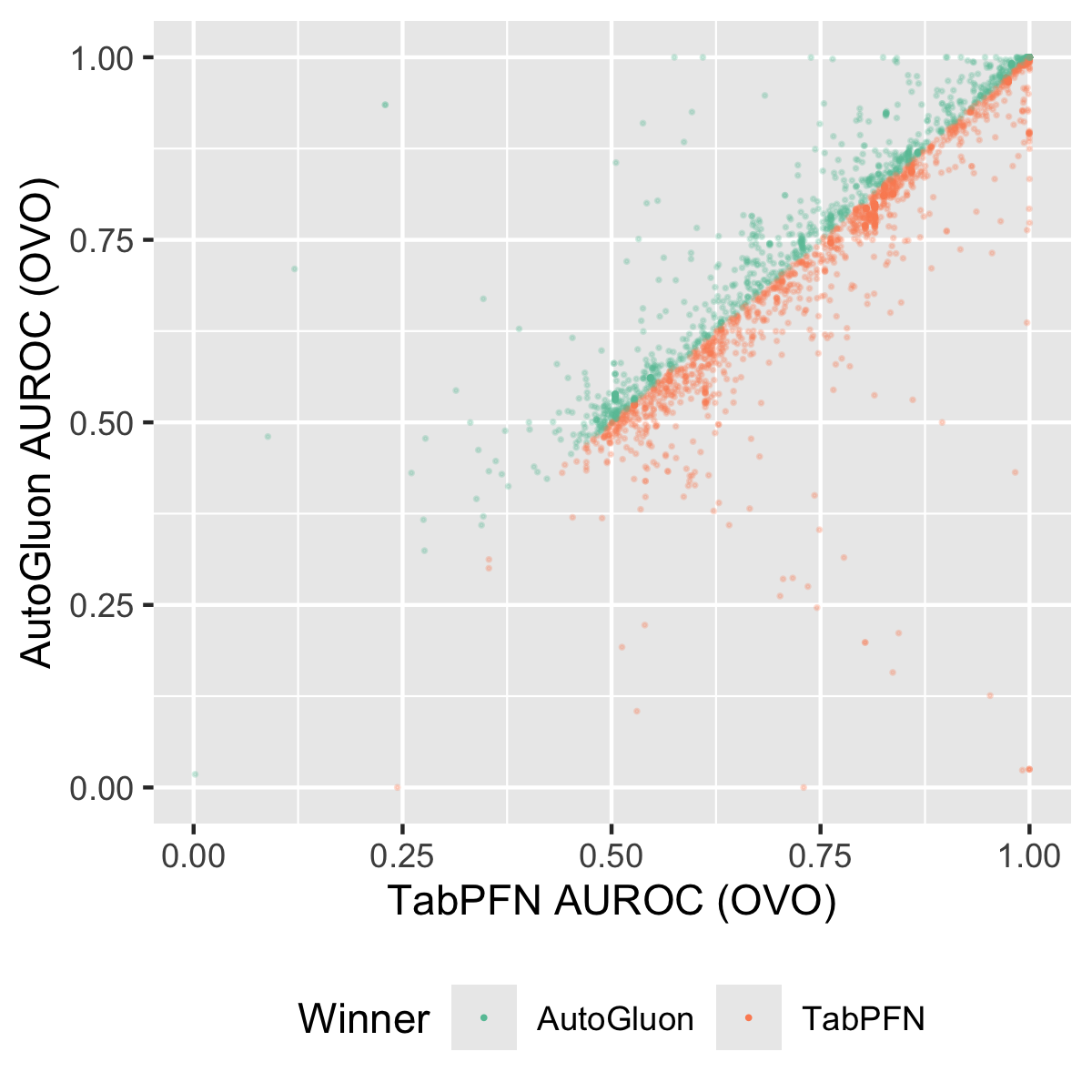}
    \vskip-8pt
    \caption{The AUROC (OVO) of TabPFN versus baseline AutoGluon with 1-minute time budget on the tabular classification problems in AnnotatedTables.
    }
    \label{fig:tabpfn-dots}
\end{wrapfigure}

\paragraph{TabPFN performs robustly on diverse tabular data} We fit TabPFN and AutoGluon on 2,720 tabular classification problems in AnnotatedTables, 
and we compute the AUROC (one-versus-one) and cross entropy for every one of them, shown in Table \ref{tab:tabpfn}.
WilCoxon signed-rank test shows that all the paired differences between TabPFN and AutoGluon for both AUROC and cross entropy and for both 1 minute and 5 minutes are non-zero with statistical significance (p-values and discussions in Appendix \ref{appx:wilcox}). 
AnnotatedTables allows us to conduct a large number of experiments to draw statistically significant conclusions,  
but the paired differences are small and may not be informative for model selection, 
an issue known generally as ``statistical significance versus practical significance'' \citep{doi:10.1177/0013164496056005002, peeters2016practical, mcshane2022selecting}. 
Overall, TabPFN performs on par with the AutoGluon baseline.

The classification problems in AnnotatedTables are diverse and have lots of inherent variations, which is reflected by the standard deviation of both models' performance. Between the two, TabPFN has a lower standard deviation for both AUROC and cross entropy, showing a more consistent performance. 

The inherent diversity of AnnotatedTables can be seen from the dot plots in Figure \ref{fig:tabpfn-dots}, where we plot the AUROC of both models versus each other. Given a relatively high spread of AUROC and overall on-par performance, it is not possible to pre-determine which model will perform better, and either model could be better on a specific table.
Practitioners may fit both models and select the better performing one in practice. Given that AutoGluon is a widely used AutoML framework, we believe that the newly introduced TabPFN is a welcome addition that would raise the state-of-the-art performance on many real-world tabular classification problems, especially as it takes only a fraction of the time budget per table
(2 seconds) compared to AutoGluon (1 or 5 minutes in our experiments).

\paragraph{LLM annotation quality}
Running TabPFN and AutoGluon on LLM-generated annotation proves that LLMs can successfully identify potential input and target columns in a table and annotate them, given only a prompt that describes each table's name, column names and data types. As seen in Figure \ref{fig:tabpfn-dots}, most of the tabular classification problems annotated by LLMs have an AUROC over 0.5 for both TabPFN and AutoGluon. This means, for most tables, the target columns identified by LLMs do have non-trivial relationships with the input columns and can be predicted to some extent, and the LLM successfully found the suitable tables for the tabular classification task.

\section{Conclusions and Discussions}

Prior to this work, human labor had been necessary to acquire tabular dataset annotations.
Large language models can bring a radical change to how machine learning datasets are constructed, and AnnotatedTables is the first to pioneer tabular dataset annotation with LLMs. %
LLMs can be steered to generate annotations based on detailed instructions, and tabular machine learning researchers could easily acquire tailored annotations that fit their research directions.
With the human annotation bottleneck opened, our methodology could enable more studies with reduced costs and automation.

\bibliography{my}
\bibliographystyle{iclr2024_conference}

\appendix

\section{Example Prompts for Large Language Models Annotation }
\label{appx:prompt}
\label{appx:sql_rel_examples}

An example of the prompt for the zero-shot synthesis of SQL code is shown below.

\lstset{
  basicstyle=\small\ttfamily,
  breaklines=true,
  breakatwhitespace=false,
  columns=fullflexible,
  frame=single,
  keepspaces=true,
  numbers=left,
  numberstyle=\tiny\color{gray},
  stepnumber=1,
  numbersep=5pt,
  moredelim=**[is][\color{red}]{@}{@},
}
\begin{lstlisting}
We have a SQL database with 2 tables. For each table, here is the schema and 
a sample row data.
Schema for table 0:
CREATE TABLE MaLga12345(
    Saledate  TEXT,
    MA  BIGINT,
    Type  TEXT,
    Bedrooms  BIGINT
);
A sample row data from table 0:
{"Saledate": "30/09/2007", "MA": 441854, "Type": "house", "Bedrooms": 2}
Schema for table 1:
CREATE TABLE RawSales(
    Datesold  TEXT,
    Postcode  BIGINT,
    Price  BIGINT,
    PropertyType  TEXT,
    Bedrooms  BIGINT
);
A sample row data from table 1:
{"Datesold": "2007-02-07 00:00:00", "Postcode": 2607, "Price": 525000, 
"PropertyType": "house", "Bedrooms": 4}
Consider the typical users who access this database. What kind of SQLite 
SELECT queries would they write? List 15 examples SQL code that are 
directly executable with relatively high complexity, each following a 
description. If possible, use joins. Reply in the format of a description 
and the SQL code for each example.

\end{lstlisting}

An example of the prompt for SQL to Rel translation is shown below.

\begin{lstlisting}
REL is a database management system language that is similar to datalog. REL is based on the sixth normal form (6NF),
where every variable represents a relation, i.e. a set of tuples.
In REL, from-variables will not appear in the output, and for-variables will appear in the output.
Strings in REL use double quotes.
Below are few examples of SQL code and REL code pairs that perform the same query.
Examples:
    
1.SQL: 
```
SELECT L.JobTitle, L.ExperienceLevel, V3.JobTitle, V3.ExperienceLevel FROM LatestDataScienceSalaries AS L INNER JOIN V3LatestDataScienceSalaries AS V3 ON L.CompanyLocation = V3.CompanyLocation WHERE L.CompanyLocation = "United States";
```
1.REL: 
```
def temp[x,y] = LatestDataScienceSalaries:CompanyLocation[x, company_location] and V3LatestDataScienceSalaries:CompanyLocation[y, company_location] and company_location = "United States" from company_location
def output[x_job_title, x_experience_level, y_job_title, y_experience_level]  = LatestDataScienceSalaries:JobTitle[x, x_job_title],  LatestDataScienceSalaries:ExperienceLevel[x, x_experience_level], V3LatestDataScienceSalaries:JobTitle[y, y_job_title], V3LatestDataScienceSalaries:ExperienceLevel[y, y_experience_level] and temp[x,y] for x, y 
```
2.SQL: 
```
SELECT p.ImageLink FROM Photos p JOIN Likes l ON p.Id = l.Photo JOIN Users u ON l.User = u.Id LEFT JOIN Follows f ON u.Id = f.Follower AND f.Followee = p.UserID WHERE f.Follower IS NULL;
```
2.REL: 
```
def UserPhoto(user_id, photo_id) = Photos:UserID[l,user_id] and Photos:Id[l, photo_id] from l
def UserLikesPhoto(user_id, photo_id) = Likes:User[l,user_id] and Likes:Photo[l, photo_id] from l
def UserID(user_id) = Users:Id(l, user_id) from l
def UserFollowsUser(follower_id, followee_id) = Follows:Follower[l,follower_id] and Follows:Followee[l, followee_id] from l
def PhotoIDHasImageLink(photo_id, image_link) =
    Photos:Id(p, photo_id) and Photos:ImageLink(p, image_link) from p
def photos_liked_by_not_a_follower(photo) =
    UserPhoto(poster, photo)
    and UserLikesPhoto(liker, photo)
    and not UserFollowsUser(liker, poster)
    from poster, liker
def output = PhotoIDHasImageLink[photos_liked_by_not_a_follower]
```
3.SQL: 
```
SELECT MainTable.StateB, MainTable.Name FROM MainTable JOIN CaseUniverse ON MainTable.StateB = CaseUniverse.TargetState AND MainTable.StateBCode = CaseUniverse.TargetCOWID WHERE TargetState = "FRN" AND TargetCOWID = "220";
```
3.REL: 
```
def join[mt, cu] = MainTable:StateB[mt, sb] and CaseUniverse:TargetState[cu, sb] and MainTable:StateBCode[mt, sbc] and CaseUniverse:TargetCOWID[cu, sbc] and sb="FRN" and sbc="220" from sb, sbc 
def output = MainTable:StateB[mt], MainTable:Name[mt] for mt, cu where join[mt, cu]
```
4.SQL: 
```
SELECT LD.Regions, SUM(DT.Punjab + DT.Haryana + DT.Rajasthan + DT.Delhi + DT.UP + DT.Uttarakhand) AS Total_Usage FROM DatasetTk DT INNER JOIN LongData LD ON DT.Punjab = LD.Usage AND LD.States = 'Punjab' GROUP BY LD.Regions; 
```
4.REL: 
```
def join1[dt,ld]= DatasetTk:Punjab[dt, punjab] and LongData:Usage[ld, punjab] from punjab
def sum_usage = DatasetTk:Punjab[dt] + DatasetTk:Haryana[dt] +
DatasetTk:Rajasthan[dt] + DatasetTk:Delhi[dt] +
DatasetTk:UP[dt] + DatasetTk:Uttarakhand[dt]  for dt
def output[region] = sum[dt, usage: 
LongData:Regions[ld, region] 
and sum_usage[dt, usage]
and join1[dt,ld] and LongData:States[ld] = "Punjab" for ld]
                
```
5.SQL: 
```
SELECT Test.ID, Test.Age, Train.SpendingScore FROM Test INNER JOIN Train ON Test.ID = Train.ID WHERE Test.Age >= 40 OR Train.Age >= 40; 
```
5.REL: 
```
def output[id, age, spending_score] = Test:ID[test_idx, id], Test:Age[test_idx, age], Train:SpendingScore[train_idx, spending_score] and  Test:ID[test_idx, id] and Train:ID[train_idx, id] and (Test:Age[test_idx]>=40 or Train:Age[train_idx] >= 40) from test_idx, train_idx
```
6.SQL: 
```
SELECT bb.PlayerOut, m.WinningTeam FROM IPLBallByBall2022 bb JOIN IPLMatches2022 m ON bb.ID = m.ID WHERE bb.ID = 1312200 AND bb.IsWicketDelivery = 1; 
```
6.REL: 
```
def join[bb, m] = IPLBallByBall2022:ID[bb, id] and IPLMatches2022:ID[m, id] and IPLBallByBall2022:ID[bb]=1312200 and IPLBallByBall2022:IsWicketDelivery[bb]=1 from id 
def output = IPLBallByBall2022:PlayerOut[bb], IPLMatches2022:WinningTeam[m] from bb, m where join[bb,m] 
```
7.SQL: 
```
SELECT il.AccountName, il.Ismarketplace FROM InstAccLabeled AS il JOIN InstAcc AS ia ON il.AccountName = ia.Username WHERE ia.IsBusinessAccount = 1;
```
7.REL: 
```
def join[il, ia] = InstAccLabeled:AccountName[il, account_name] and InstAcc:Username[ia, account_name] and InstAcc:IsBusinessAccount[ia]=boolean_true from account_name 
def output = InstAccLabeled:AccountName[il], InstAccLabeled:Ismarketplace[il]  for il, ia where join[il, ia]
```
8.SQL: 
```
SELECT table0.Area, table0.AveragePrice, table1.MedianSalary FROM HousingInLondonMonthlyVariables AS table0 JOIN HousingInLondonYearlyVariables AS table1 ON table0.Code = table1.Code WHERE table0.BoroughFlag = 1;
```
8.REL: 
```
def join[row_id_0, row_id_1] = HousingInLondonMonthlyVariables:Code[row_id_0, code] and HousingInLondonYearlyVariables:Code[row_id_1, code] and HousingInLondonMonthlyVariables:BoroughFlag[row_id_0]=1 from code   
def output[area, average_price, median_salary] = HousingInLondonMonthlyVariables:Area[row_id_0, area] and HousingInLondonMonthlyVariables:AveragePrice[row_id_0, average_price] and HousingInLondonYearlyVariables:MedianSalary[row_id_1, median_salary] and join[row_id_0, row_id_1] for row_id_0, row_id_1  
```
9.SQL: 
```
SELECT Gender, Segmentation, COUNT(*) FROM Test WHERE Segmentation IN ('B', 'C') GROUP BY Gender, Segmentation;
```
9.REL: 
```
def output[gender, seg] = count[row_id: Test:Gender(row_id, gender) and Test:Segmentation(row_id,seg) and {"B"; "C"}(seg)] 
```
10.SQL: 
```
SELECT DISTINCT State FROM NCHSLeadingCausesOfDeathUnitedStates; 
```
10.REL: 
```
def output[state] = NCHSLeadingCausesOfDeathUnitedStates:State[x,state] from x
```
11.SQL: 
```
SELECT * FROM HRCommaSep;
```
11.REL: 
```
def output[col, row, val] = HRCommaSep[col, row, val]
```
12.SQL: 
```
SELECT Title, Year FROM TvShows;
```
12.REL: 
```
def output[title, year] = TvShows:Title[idx, title], TvShows:Year[idx, year] for idx
```
13.SQL: 
```
SELECT AverageMontlyHours FROM HRCommaSep WHERE Department = 'sales';
```
13.REL: 
```
def output[average_monthly_hours] = HRCommaSep:Department(row_idx, "sales") and HRCommaSep:AverageMontlyHours(row_idx, average_monthly_hours) for row_idx
```
14.SQL: 
```
SELECT Location, Age FROM User WHERE UserId = 1;
```
14.REL: 
```
def output[location, age] = User:Location[idx, location], User:Age[idx, age], User:UserId[idx, 1] for idx 
```
15.SQL: 
```
SELECT * FROM GMPlayersStatistics WHERE isstreamer = 1 AND bulletwin > 70;
```
15.REL: 
```
def output[col, row, val] = GMPlayersStatistics[col, row, val] and GMPlayersStatistics:IsStreamer[row] = boolean_true and GMPlayersStatistics:BulletWin[row] > 70  
```
16.SQL: 
```
SELECT AVG(Visibilitykm), MAX(WindSpeedkmh) FROM WeatherData WHERE Weather = "Rain";
```
16.REL: 
```
def all_row_ids_with_rain[row_id] = WeatherData:Weather[row_id]="Rain" 
def output = average[row_id, value_vis: all_row_ids_with_rain(row_id) and WeatherData:VisibilityKm(row_id, value_vis)], max[row_id, value_wind_speed : all_row_ids_with_rain(row_id) and WeatherData:WindSpeedKmh(row_id, value_wind_speed)] 
```
17.SQL: 
```
SELECT * FROM Data WHERE Danceability > 0.8;
```
17.REL: 
```
def output[col, idx, val] = Data[col, idx, val] and Data:Danceability[idx] > 0.8
```
18.SQL: 
```
SELECT * FROM SpaceCorrected WHERE StatusMission IN ('Success', 'Failure');
```
18.REL: 
```
def output[col, row_id, val] = SpaceCorrected[col, row_id, val] and {"Success"; "Failure"}(SpaceCorrected:StatusMission[row_id])
```
19.SQL: 
```
SELECT * FROM SpaceCorrected WHERE Location LIKE '%
```
19.REL: 
```
def output[col, row_id, val] = SpaceCorrected[col, row_id, val] and like_match("\%Florida\%", SpaceCorrected:Location[row_id])
```
20.SQL: 
```
SELECT * FROM Cars24Combined WHERE Fuel = 'PETROL';
```
20.REL: 
```
def output[colname, row_idx, val] = Cars24Combined[colname, row_idx, val] and Cars24Combined[:Fuel, row_idx, "PETROL"]
```
21.SQL: 
```
SELECT DISTINCT Name FROM UCIDatasets;
```
21.REL: 
```
def output[name] = UCIDatasets:Name[idx, name] from idx 
```
22.SQL: 
```
SELECT COUNT(*) FROM SupermarketSalesSheet1;
```
22.REL: 
```
def count_rows = count[row_id: SupermarketSalesSheet1[_, row_id, _]]
 def output = count_rows
```
23.SQL: 
```
SELECT * FROM Jobs WHERE ActivelyHiring = 1.0;
```
23.REL: 
```
def output[col, idx, val] = Jobs[col, idx, val] and Jobs:ActivelyHiring[idx] = 1.0
```
24.SQL: 
```
SELECT marvelcomiccharactername, comicappearance1 FROM MarvelComicsLegacy WHERE marvelcomiccharactername = 'Wolverine';
```
24.REL: 
```
def output[marvel_character_name, comic_appearance_1] = MarvelComicsLegacy:MarvelComicCharacterName[idx, marvel_character_name], MarvelComicsLegacy:ComicAppearance1[idx, comic_appearance_1] and marvel_character_name = "Wolverine" for idx
```
25.SQL: 
```
SELECT * FROM WAFnUseCTelcoCustomerChurn WHERE Churn = "Yes";
```
25.REL: 
```
def output[col, row_idx, val] = WAFnUseCTelcoCustomerChurn[col, row_idx, val] and WAFnUseCTelcoCustomerChurn:Churn[row_idx] = "Yes"
```
26.SQL: 
```
SELECT AVG(Charges) FROM Insurance WHERE Age > 40;
```
26.REL: 
```
def output = average[idx, charges: Insurance:Charges[idx, charges] and Insurance:Age[idx] > 40]
```
27.SQL: 
```
SELECT * FROM CommentsCleaned WHERE EmojiUsed = "yes";
```
27.REL: 
```
def output[col, row_id, val] = CommentsCleaned[col, row_id, val] and CommentsCleaned:EmojiUsed[row_id] = "yes"
```
28.SQL: 
```
SELECT * FROM Cars24Combined WHERE Fuel = "PETROL" AND Location = "HR-98";
```
28.REL: 
```
def output[col, idx, val] = Cars24Combined[col, idx, val] and Cars24Combined:Fuel[idx, "PETROL"] and Cars24Combined:Location[idx, "HR-98"]
```
29.SQL: 
```
SELECT StateUT, AVG(NumberOfHouseholdsSurveyed) as AverageHouseholds FROM Datafile GROUP BY StateUT;
```
29.REL: 
```
def output[state] = average[idx, num: Datafile:NumberOfHouseholdsSurveyed[idx, num] and Datafile:StateUT[idx, state]]
```
30.SQL: 
```
SELECT * FROM Cars24Combined WHERE Fuel = 'PETROL' AND Location = 'HR-98';
```
30.REL: 
```
def output[col, idx, val] = Cars24Combined[col, idx, val] and Cars24Combined:Fuel[idx, "PETROL"] and Cars24Combined:Location[idx, "HR-98"]
```
31.SQL: 
```
SELECT Gender, COUNT(*) FROM Diabetes GROUP BY Gender;
```
31.REL: 
```
def output[gender] = count[row_id: Diabetes:Gender[row_id, gender]]
```
32.SQL: 
```
SELECT JobTitle, AVG(Salary) AS AverageSalary
FROM PartiallyCleanedSalaryDataset
GROUP BY JobTitle;
```
32.REL: 
```
def output[job_title] = average[idx, salary: PartiallyCleanedSalaryDataset:Salary[idx, salary] and PartiallyCleanedSalaryDataset:JobTitle[idx, job_title]]
```
33.SQL: 
```
SELECT Race, COUNT(*) AS TotalPitstops
FROM Pitstops
GROUP BY Race;
```
33.REL: 
```
def output[race] = count[row_id: Pitstops:Race[row_id, race]]
```
34.SQL: 
```
SELECT Title, Rating FROM TopAnime WHERE Rating = (SELECT MAX(Rating) FROM TopAnime);
```
34.REL: 
```
def max_rating = max[x, rating : TopAnime:Rating[x, rating]] 
def output[title, rating] = TopAnime:Title[idx, title], TopAnime:Rating[idx, rating] and TopAnime:Rating[idx, max_rating] for idx
```
35.SQL: 
```
SELECT Category, SUM(Sales) AS TotalSales, SUM(Profit) AS TotalProfit 
FROM SampleSuperstore 
GROUP BY Category;
```
35.REL: 
```
def output[category] = sum[row_id, sales: SampleSuperstore:Category(row_id, category) and SampleSuperstore:Sales(row_id, sales)], sum[row_id, profit: SampleSuperstore:Category(row_id, category) and SampleSuperstore:Profit(row_id, profit)]
```
36.SQL: 
```
SELECT SUM(CustomerCareCalls) 
FROM Train 
WHERE DiscountOffered >= 50;
```
36.REL: 
```
def output = sum[row_id,calls: Train:CustomerCareCalls[row_id, calls] and Train:DiscountOffered[row_id] >= 50]
```
37.SQL: 
```
SELECT AVG(MonthlyCharges) FROM CustomerChurn WHERE Churn = 'Yes';
```
37.REL: 
```
def output = average[row_id, monthly_charges: CustomerChurn:MonthlyCharges(row_id, monthly_charges) and CustomerChurn:Churn[row_id] = "Yes"]
```
38.SQL: 
```
SELECT COUNT(DISTINCT StatusType) AS UniqueStatusTypes
FROM Live;
```
38.REL: 
```
def output = count[status_type: Live:StatusType(row_id, status_type) from row_id]
```
39.SQL: 
```
SELECT * FROM PCOSInfertility ORDER BY SlNo DESC;
```
39.REL: 
```
def output[col, row_idx, val] = PCOSInfertility[col, row_idx, val]
```
40.SQL: 
```
SELECT SUM(WeeklySales) AS TotalWeeklySales FROM WalmartStoreSales;
```
40.REL: 
```
def output = sum[row_idx, weekly_sales : WalmartStoreSales:WeeklySales[row_idx, weekly_sales]]
```
41.SQL: 
```
SELECT * FROM WeatherData ORDER BY RelativeHumidity ASC;
```
41.REL: 
```
def output[col, row_id, val] = WeatherData[col, row_id, val]
```
42.SQL: 
```
SELECT ShipMode, SUM(Sales) AS TotalSales FROM SampleSuperstore GROUP BY ShipMode;
```
42.REL: 
```
def output[shipmode] = sum[row_id, value_sales: SampleSuperstore:Sales[row_id, value_sales] and SampleSuperstore:ShipMode[row_id, shipmode]]
```
43.SQL: 
```
SELECT PlayerOfMatch, WinningTeam
FROM IPLMatches2022
WHERE WinningTeam = "Rajasthan Royals";
```
43.REL: 
```
def output[player_of_match, winning_team] = IPLMatches2022:PlayerOfMatch[idx, player_of_match], IPLMatches2022:WinningTeam[idx, winning_team] and IPLMatches2022:WinningTeam[idx, winning_team] and winning_team = "Rajasthan Royals" for idx
```
44.SQL: 
```
SELECT Title, Desc FROM MegaGymDataset WHERE Type = "Strength";
```
44.REL: 
```
def output[title, desc] = MegaGymDataset:Title[idx, title], MegaGymDataset:Desc[idx, desc]  and MegaGymDataset:Type[idx, "Strength"] for idx
```
45.SQL: 
```
SELECT TypeOfInternship, COUNT(*) as Count FROM Internship GROUP BY TypeOfInternship;
```
45.REL: 
```
def output[type_of_internship] = count[row_id: Internship:TypeOfInternship[row_id, type_of_internship]]
```
Given examples above, translate the following SQL queries to REL programs. No explanation is needed.
Output only the REL code one by one numbered with '1.REL:', '2.REL:', for example.
    
1.SQL:
```
SELECT * FROM BookingSaudiArabia ORDER BY Score DESC LIMIT 5;
```
2.SQL:
```
SELECT * 
FROM VsrrProvisionalDrugOverdoseDeathCounts
WHERE DataValue > 5000;
```
3.SQL:
```
SELECT COUNT(*) FROM Tweets;
```
4.SQL:
```
SELECT AVG(PredSvr) FROM PredSvr INNER JOIN PredXgb ON PredSvr.PredSvr = PredXgb.XgCost;
```
5.SQL:
```
SELECT * FROM ChurnTrainTransformed WHERE CreditPerProduct < 400 OR EstSalaryScaled IS NULL
UNION
SELECT * FROM TestDfTransformed WHERE CreditPerProduct < 400 OR EstSalaryScaled IS NULL;
```
6.SQL:
```
SELECT * FROM Leads WHERE LastNotableActivity = 'Modified';
```
7.SQL:
```
SELECT CentralPressure, MaximumSustainedWindSpeed FROM TyphoonData WHERE IndicatorOfLandfallOrPassage = 'Landfall';
```
8.SQL:
```
SELECT SUM(A28) FROM Texture WHERE A17 < (SELECT AVG(A17) FROM Texture);
```
9.SQL:
```
SELECT FlatModel, MAX(ResalePrice) AS MaxResalePrice
    FROM ResaleFlatPricesBasedOnApprovalDate2000Feb2012
    GROUP BY FlatModel;
```
10.SQL:
```
SELECT COUNT(*) FROM TestFile;
```
\end{lstlisting}

An example of the prompt for input-target column annotation is shown below.

\begin{lstlisting}
Consider a machine learning model that takes a few numeric input columns 
and predict a single classification target column. Given the following
schema of a data table, suggest the input columns and target column, such
that the target may be predicted from the inputs non-trivially.
Schema: CREATE TABLE PartiallyCleanedSalaryDataset(
    CompanyName  TEXT,
    JobTitle  TEXT,
    SalariesReported  BIGINT,
    Location  TEXT,
    Salary  FLOAT
);
Respond in JSON format with `input_columns' and `output_column'. 
\end{lstlisting}

\section{Steering LLMs to Generate More Complex SQL Programs}
\label{appx:sql_steering_complex}
During prototyping, we refined the instructions to create the best annotation results. One of our goals was to increase the complexity of SQL programs generated by the LLM. We experimented with different prompts, and we found that instructing the LLM to generate SQL programs ``with relatively high complexity'' can increase the complexity and reduce the number of simple queries generated by the LLM. An example is given here to illustrate the effect of including the steering phrase and without the steering phrase.

The following is a prompt for the zero-shot synthesis of SQL code without the steering phrase ``with relatively high complexity''.

\begin{lstlisting}
We have a SQL database with 2 tables. For each table, here is the schema and 
a sample row data.
Schema for table 0:
CREATE TABLE MaLga12345(
    Saledate  TEXT,
    MA  BIGINT,
    Type  TEXT,
    Bedrooms  BIGINT
);
A sample row data from table 0:
{"Saledate": "30/09/2007", "MA": 441854, "Type": "house", "Bedrooms": 2}
Schema for table 1:
CREATE TABLE RawSales(
    Datesold  TEXT,
    Postcode  BIGINT,
    Price  BIGINT,
    PropertyType  TEXT,
    Bedrooms  BIGINT
);
A sample row data from table 1:
{"Datesold": "2007-02-07 00:00:00", "Postcode": 2607, "Price": 525000, 
"PropertyType": "house", "Bedrooms": 4}
Consider the typical users who access this database. What kind of SQLite 
SELECT queries would they write? @List 15 examples SQL code that are 
directly executable, each following a description.@ If possible, use joins. 
Reply in the format of a description and the SQL code for each example.
\end{lstlisting}

Here is the annotation generated by the LLM when the steering phrase is not used.
\begin{lstlisting}
1. Get all rows from MaLga12345 table:
```sql
SELECT * FROM MaLga12345;
```

2. Get all rows from RawSales table:
```sql
SELECT * FROM RawSales;
```

3. Find the average price of properties sold by postcode:
```sql
SELECT Postcode, AVG(Price) as AveragePrice
FROM RawSales
GROUP BY Postcode;
```

4. List all house sales from MaLga12345:
```sql
SELECT * FROM MaLga12345
WHERE Type = 'house';
```

5. List all properties sold on a specific date from RawSales:
```sql
SELECT * FROM RawSales
WHERE Datesold = '2007-02-07 00:00:00';
```

6. Find the total number of bedrooms sold in RawSales:
```sql
SELECT SUM(Bedrooms) as TotalBedrooms
FROM RawSales;
```

7. Get the maximum MA value from MaLga12345:
```sql
SELECT MAX(MA) as MaxMA
FROM MaLga12345;
```

8. List properties sold in a specific postcode with more than 3 bedrooms from RawSales:
```sql
SELECT * FROM RawSales
WHERE Postcode = 2607 AND Bedrooms > 3;
```

9. Count the number of sales by property type from RawSales:
```sql
SELECT PropertyType, COUNT(*) as SalesCount
FROM RawSales
GROUP BY PropertyType;
```

10. Get all properties sold in a specific year from RawSales:
```sql
SELECT * FROM RawSales
WHERE Datesold LIKE '2007%
```

11. Find the sum of prices for all house sales from RawSales:
```sql
SELECT SUM(Price) as TotalHouseSales
FROM RawSales
WHERE PropertyType = 'house';
```

12. Join MaLga12345 and RawSales tables on bedrooms and list matching records:
```sql
SELECT a.*, b.*
FROM MaLga12345 a
JOIN RawSales b ON a.Bedrooms = b.Bedrooms;
```

13. Find properties with the same sale date and date sold from both tables:
```sql
SELECT a.*, b.*
FROM MaLga12345 a
JOIN RawSales b ON a.Saledate = b.Datesold;
```

14. Get the count of properties sold for each MA value from MaLga12345:
```sql
SELECT MA, COUNT(*) as CountOfSales
FROM MaLga12345
GROUP BY MA;
```

15. List all properties from MaLga12345 and their corresponding prices from RawSales if they have the same number of bedrooms:
```sql
SELECT a.*, b.Price
FROM MaLga12345 a
JOIN RawSales b ON a.Bedrooms = b.Bedrooms;
```
\end{lstlisting}

The following is a prompt for the zero-shot synthesis of SQL code that has the steering phrase ``with relatively high complexity''.

\begin{lstlisting}
We have a SQL database with 2 tables. For each table, here is the schema and 
a sample row data.
Schema for table 0:
CREATE TABLE MaLga12345(
    Saledate  TEXT,
    MA  BIGINT,
    Type  TEXT,
    Bedrooms  BIGINT
);
A sample row data from table 0:
{"Saledate": "30/09/2007", "MA": 441854, "Type": "house", "Bedrooms": 2}
Schema for table 1:
CREATE TABLE RawSales(
    Datesold  TEXT,
    Postcode  BIGINT,
    Price  BIGINT,
    PropertyType  TEXT,
    Bedrooms  BIGINT
);
A sample row data from table 1:
{"Datesold": "2007-02-07 00:00:00", "Postcode": 2607, "Price": 525000, 
"PropertyType": "house", "Bedrooms": 4}
Consider the typical users who access this database. What kind of SQLite 
SELECT queries would they write? List 15 examples SQL code that are 
directly executable @with relatively high complexity@, each following a 
description. If possible, use joins. Reply in the format of a description 
and the SQL code for each example.
\end{lstlisting}

The following is the annotation generated by the LLM when the steering phrase is used. Clearly, the SQL programs generated have a higher complexity, and they are more valuable as SQL annotations for our data tables. 

\begin{lstlisting}
1. Select all sales records from both tables where the property type is a "house":
```sql
SELECT * FROM MaLga12345 WHERE Type = 'house'
UNION
SELECT * FROM RawSales WHERE PropertyType = 'house';
```

2. Retrieve the total price for houses sold in table 1:
```sql
SELECT SUM(Price) as total_price FROM RawSales WHERE PropertyType = 'house';
```

3. List the average number of bedrooms for properties sold in each table:
```sql
SELECT AVG(Bedrooms) as avg_bedrooms_table0 FROM MaLga12345;
SELECT AVG(Bedrooms) as avg_bedrooms_table1 FROM RawSales;
```

4. Find the number of sales made in each table in the year 2007:
```sql
SELECT COUNT(*) as sales_count_table0 FROM MaLga12345 WHERE Saledate LIKE '%
SELECT COUNT(*) as sales_count_table1 FROM RawSales WHERE Datesold LIKE '%
```

5. Retrieve the sales records where the postcode is 2607 from table 1:
```sql
SELECT * FROM RawSales WHERE Postcode = 2607;
```

6. Identify the latest sale date recorded from both tables:
```sql
SELECT MAX(Saledate) as latest_sale_date_table0 FROM MaLga12345;
SELECT MAX(Datesold) as latest_sale_date_table1 FROM RawSales;
```

7. Display the details for sales records with a price greater than $500,000 in table 1:
```sql
SELECT * FROM RawSales WHERE Price > 500000;
```

8. Retrieve the number of bedrooms for properties sold in table 0, grouped by property type:
```sql
SELECT Type, AVG(Bedrooms) as avg_bedrooms FROM MaLga12345 GROUP BY Type;
```

9. List the sales records from both tables where the number of bedrooms is greater than 3:
```sql
SELECT * FROM MaLga12345 WHERE Bedrooms > 3
UNION
SELECT * FROM RawSales WHERE Bedrooms > 3;
```

10. Calculate the average price of properties sold in table 1 with 4 bedrooms:
```sql
SELECT AVG(Price) as avg_price FROM RawSales WHERE Bedrooms = 4;
```

11. Count the number of distinct postcodes present in table 1:
```sql
SELECT COUNT(DISTINCT Postcode) as unique_postcodes FROM RawSales;
```

12. Retrieve the sales records from table 1 where the price is greater than the average price:
```sql
SELECT * FROM RawSales 
WHERE Price > (SELECT AVG(Price) FROM RawSales);
```

13. Join tables based on matching postal codes and display the combined dataset:
```sql
SELECT * FROM MaLga12345 
JOIN RawSales ON MaLga12345.MA = RawSales.Postcode;
```

14. Identify the property types present in the two tables without duplication:
```sql
SELECT DISTINCT Type FROM MaLga12345
UNION
SELECT DISTINCT PropertyType FROM RawSales;
```

15. Show the top 5 highest-priced sales records from both tables:
```sql
SELECT * FROM (SELECT * FROM MaLga12345
UNION
SELECT * FROM RawSales)
ORDER BY Price DESC
LIMIT 5;
```
\end{lstlisting}

SQL queries with higher complexity reflect better annotation quality. Thus we include the steering phrase ``with relatively high complexity'' in our prompt.

\section{Costs of LLM Annotation}

We use the public service from OpenAI to access ChatGPT with the GPT-3.5 large language model. To support our claim that LLM annotations are cheap and accessible, we provide an estimate of our annotation cost.

\begin{enumerate}[itemsep=0.2ex,partopsep=0.2ex,parsep=.2ex, leftmargin=3ex]
    \item Generating SQL programs costs less than 400 U.S. dollars. 
    \item Generating Rel translations for SQL programs costs less than 2000 U.S. dollars. The prompt for SQL-to-Rel translation is very long, as seen in Appendix \ref{appx:prompt}.
    \item Generating input-output column annotations for the data tables costs less than 100 U.S. dollars. 
\end{enumerate}

The time spent on annotation depends on the account limitations. OpenAI API calls can be made asynchronously to parallize the annotation process. All annotation processes combined can be completed within 2 days.

\section{Wilcoxon Signed-Rank Test for TabPFN and AutoGluon Tabular Classification Performance Difference}
\label{appx:wilcox}

\begin{figure}[t]
    \centering
    \begin{subfigure}[b]{0.23\textwidth}
        \centering
        \includegraphics[width=\textwidth]{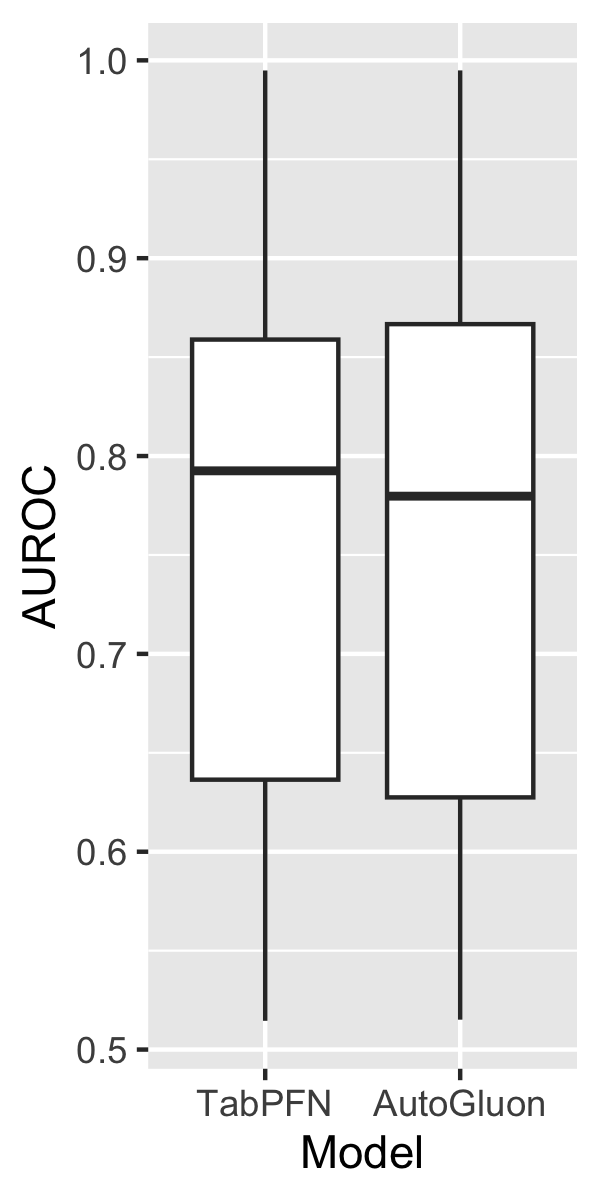}
        \caption{AUROC with 1 minute time budget.}
        \label{fig:subfig1}
    \end{subfigure}
    \hfill
    \begin{subfigure}[b]{0.23\textwidth}
        \centering
        \includegraphics[width=\textwidth]{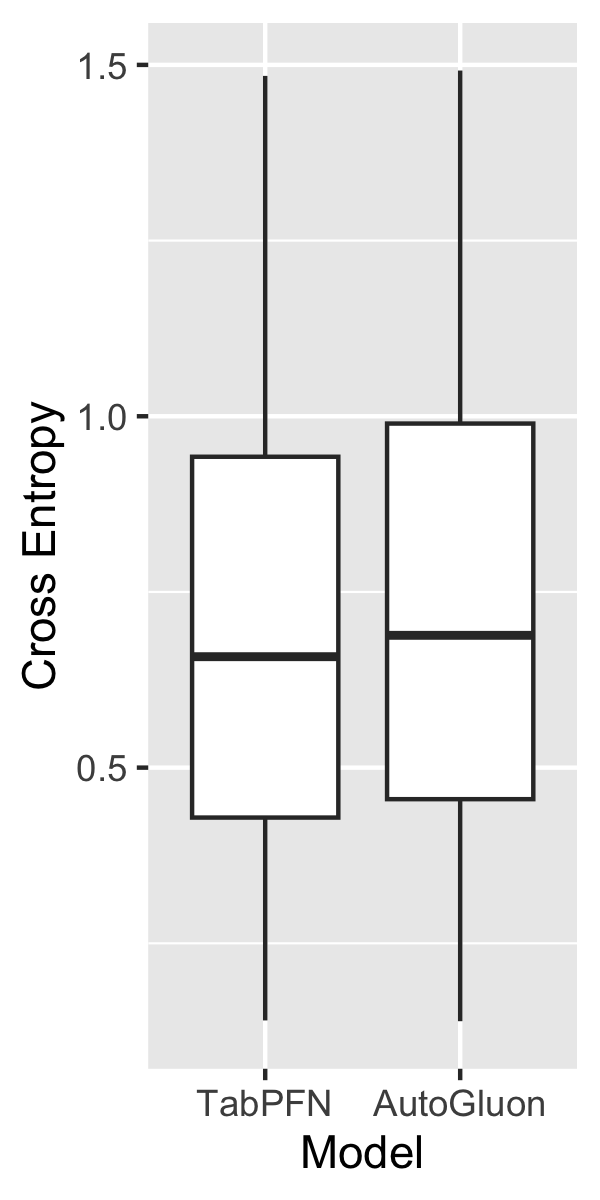}
        \caption{Cross entropy with 1 minute time budget.}
        \label{fig:subfig2}
    \end{subfigure}
    \hfill
    \begin{subfigure}[b]{0.23\textwidth}
        \centering
        \includegraphics[width=\textwidth]{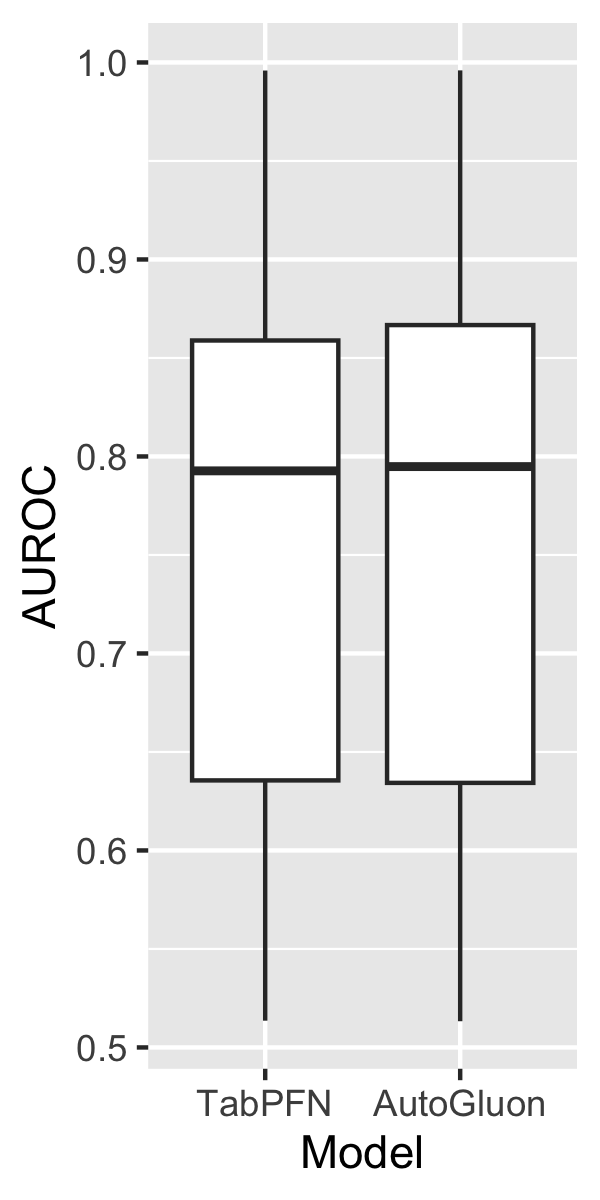}
        \caption{AUROC with 5 minutes time budget.}
        \label{fig:subfig1}
    \end{subfigure}
    \hfill
    \begin{subfigure}[b]{0.23\textwidth}
        \centering
        \includegraphics[width=\textwidth]{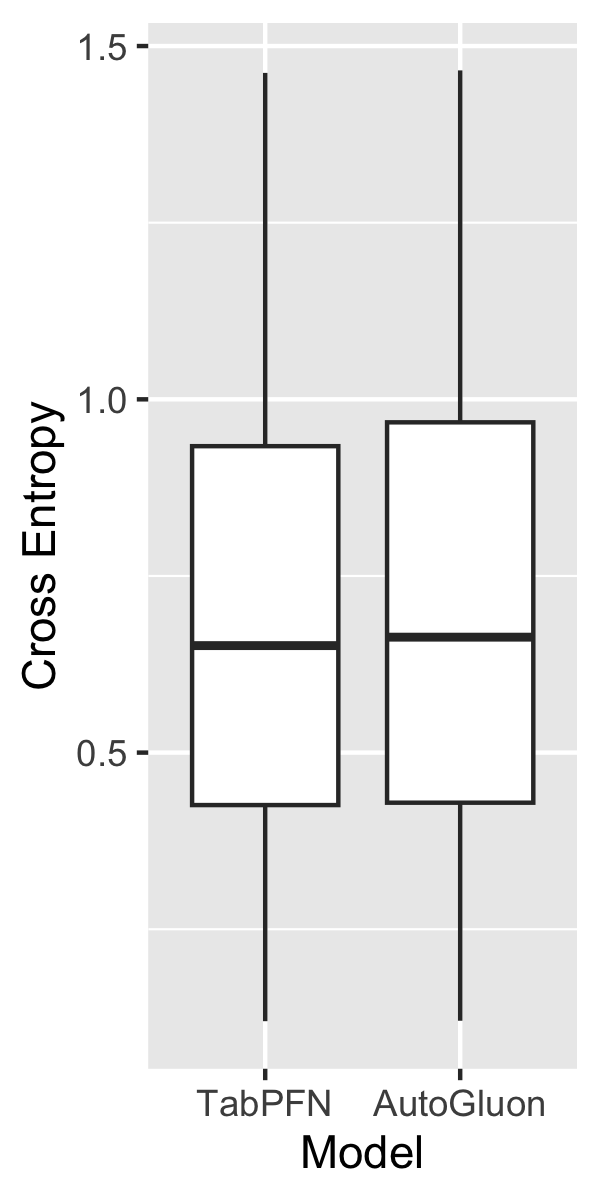}
        \caption{Cross entropy with 5 minutes time budget.}
        \label{fig:subfig2}
    \end{subfigure}
    \caption{Bar plots for the performance metrics of TabPFN and AutoGluon. The outliers are not plotted as they are far from the quantile bars. }
    \label{fig:barplot}
\end{figure}

Wilcoxon signed-rank test \citep{c4091bd3-d888-3152-8886-c284bf66a93a} is a non-parametric statistical test for paired data. We use the test to determine if there is a statistically significant performance difference between TabPFN and AutoGluon on the tabular classification problems from AnnotatedTables. For every tabular classification problem among the 2,720 problems, for every specific time budget (either 1 min or 5 mins), the performance metrics (either AUROC or cross entropy) of TabPFN and AutoGluon form a paired data. The tests show that the median difference between the pairs is nonzero with statistical significance for all settings. The p-value of Wilcoxon signed-rank test is shown in Table \ref{tab:wilcoxon}. The paired bar plots are in Figure \ref{fig:barplot}.

Judging by the results of the Wilcoxon signed-rank test, TabPFN performs better than AutoGluon with 1 minute time budget, and AutoGluon with 5 minutes time budget performs better than TabPFN. However, the difference of medians is small, and the standard deviation of performances are high.
Based solely on the general results of Wilcoxon signed-rank test, we cannot predict if one model will outperform the other for a particular tabular classification problem. The issue of ``statistical significance verus practical significance'' \citep{doi:10.1177/0013164496056005002} is reflected here: even if a result is statistically significant, it may not be practically significant if the effect size is very small.

\begin{table}[t]
\caption{The alternative hypotheses and p-values of paired Wilcoxon signed-rank test for TabPFN and AutoGluon performance on the tabular classification problems in AnnotatedTables. All alternative hypotheses are accepted with high statistical significance.
}
    \centering
\begin{tabular}{lrrrr}
\toprule
 & & 1 Min (2725 problems) & 5 Mins (2720 problems) \\
\midrule
\multirow{2}{*}{AUROC} & Alternative hypothesis & TabPFN > Autogluon & TabPFN < Autogluon\\
 & p-value &  3.798882e-20 &  6.643113e-11 \\
 \midrule
 \multirow{2}{*}{Cross entropy} & Alternative hypothesis & TabPFN < Autogluon & TabPFN > Autogluon\\
 & p-value &  4.25193e-63 & 2.44812e-06\\
\bottomrule
\end{tabular}
    \label{tab:wilcoxon}
\end{table}

\section{Data Retention Rate During Dataset Construction}
\label{appx:discard_rate}
During the dataset construction, we discard the data point if data processing fails at any step in order to maintain high quality results. In this section, we list the main data processing steps and the number of data points remaining at every step.

All intermediate data artifacts will be released as a part of the AnnotatedTables dataset. We hope future researchers can study the quality of LLM annotations and introduce methods for improvements.

\paragraph{SQL Annotation} SQL annotation starts with 70,000 Kaggle datasets and create 405,616 SQL programs with 32,119 databases, in the following data processing steps.

\begin{enumerate}[itemsep=0.2ex,partopsep=0.2ex,parsep=.2ex, leftmargin=6ex, labelsep=3ex]
    \item 70,000 Kaggle datasets are collected.
    \item 42,776 Kaggle datasets have CSV files to load as databases, representing 61.11\% of the datasets collected from the last step. 
    \item 34,460 SQL databases can be loaded, for which the schema and example row descriptions can be extracted, representing 80.56\% of the datasets with CSV files from the last step.
    \item 32,926 SQL databases have SQL annotations generated by the Large Language Model, representing 95.54\% of the datasets from the last step. The total number of SQL programs synthesized is 493,495.
    \item We execute 493,134 SQL programs from 32,902 SQL databases within a time out of 120 seconds per query to collect execution results, representing 99.93\% of the SQL programs and 99.93\% of the databases from the last step. Every table in the database is truncated to a maximum of 1,000 rows before execution of SQL programs.
    \item 405,616 SQL programs from 32,119 databases have execution results that are not empty, labeled as ``Valid'' in Table \ref{tab:sql}, representing 82.25\% of the SQL programs executed.
\end{enumerate}

\paragraph{SQL-to-Rel translation} SQL-to-Rel translation uses the SQL annotations in AnnotatedTables to translate to Rel programs.

\begin{enumerate}[itemsep=0.2ex,partopsep=0.2ex,parsep=.2ex, leftmargin=6ex, labelsep=3ex]
    \item 405,616 SQL programs from 32,119 databases are valid.
    \item 28,514 databases have at least 10 SQL programs each, forming 285,140 SQL programs to be translated, reported in Table \ref{tab:rel}, representing 70.30\% of the valid SQL programs from the last step.
    \item 116,327 Rel programs are translated correctly, with the same execution results as the SQL queries, representing 40.80\% of the source SQL programs from the last step.
\end{enumerate}

\paragraph{Input-output column annotation}
The input-output columns are annotated for the evaluation of TabPFN. For a given table, numeric input columns and a categorical output column need to be identified for the tabular classification task. For this annotation, we start with the 70,000 Kaggle datasets and evaluate TabPFN and AutoGluon on 2,720 tabular classification problems.
\begin{enumerate}[itemsep=0.2ex,partopsep=0.2ex,parsep=.2ex, leftmargin=6ex, labelsep=3ex]
    \item 70,000 Kaggle datasets are collected.
    \item 42,776 Kaggle datasets have CSV files to load as databases, representing 61.11\% of the datasets collected from the last step. 
    \item 34,009 tables are annotated with input-output columns by the LLM. One table is taken from each database to avoid re-using similar or related tables from the same database. 79.50\% of the tables from the previous step have input-output columns annotated from the LLM.
    \item 10,529 tables have numeric input columns and a categorical output column, to satisfy TabPFN's requirements. The LLM may still generate an annotation when the table does not have columns that satisfy the requirement. We do not remove missing values from the tables or tables with missing values.
    \item We run both TabPFN and AutoGluon on every table from the last step. Data is processed using TabPFN authors' code, with data truncated to at most 1,000 training samples, 100 numerical features, and 10 classes. 2,725 problems can be run and finish running within 1 minute time budget for both models, and 2,720 problems within 5 minute time budget.
\end{enumerate}

\section{Tabular Dataset Annotation and Semantic Annotation of Tabular Data}
Our paper studies the ``tabular dataset annotation problem'', that is the problem of dataset annotation limited to the tabular data domain. Based on the common concept of dataset annotation, tabular dataset annotation is the process of labeling tabular data for machine learning. Currently, tabular dataset annotation is done through expert annotators or crowd-sourced human workers.

The ``tabular dataset annotation problem'' should not be confused with ``semantic annotation of tabular data'', which is also called ``tabular data annotation'' in some context \citep{tableannotationpaperswithcode}. ``Semantic annotation of tabular data'' aims to match tabular data with knowledge graph entities.

\newpage
\section*{NeurIPS Paper Checklist}

\begin{enumerate}

\item {\bf Claims}
    \item[] Question: Do the main claims made in the abstract and introduction accurately reflect the paper's contributions and scope?
    \item[] Answer: \answerYes{} %
    \item[] Justification: The abstract and introductions made claims that are accurate and supported by the results in the paper. Our dataset is the largest SQL dataset with tabular data that support execution in the literature. Our paper is also the first to use large language models to annotate tabular datasets, to the best of our knowledge. %
    \item[] Guidelines:
    \begin{itemize}
        \item The answer NA means that the abstract and introduction do not include the claims made in the paper.
        \item The abstract and/or introduction should clearly state the claims made, including the contributions made in the paper and important assumptions and limitations. A No or NA answer to this question will not be perceived well by the reviewers. 
        \item The claims made should match theoretical and experimental results, and reflect how much the results can be expected to generalize to other settings. 
        \item It is fine to include aspirational goals as motivation as long as it is clear that these goals are not attained by the paper. 
    \end{itemize}

\item {\bf Limitations}
    \item[] Question: Does the paper discuss the limitations of the work performed by the authors?
    \item[] Answer:\answerYes{} %
    \item[] Justification: We measure and report the performance of LLM annotations, including statistics and analysis on the erroneous annotations throughout this paper. We have paragraphs in the paper devoted to discussing the limitations of our results. 
    \item[] Guidelines:
    \begin{itemize}
        \item The answer NA means that the paper has no limitation while the answer No means that the paper has limitations, but those are not discussed in the paper. 
        \item The authors are encouraged to create a separate "Limitations" section in their paper.
        \item The paper should point out any strong assumptions and how robust the results are to violations of these assumptions (e.g., independence assumptions, noiseless settings, model well-specification, asymptotic approximations only holding locally). The authors should reflect on how these assumptions might be violated in practice and what the implications would be.
        \item The authors should reflect on the scope of the claims made, e.g., if the approach was only tested on a few datasets or with a few runs. In general, empirical results often depend on implicit assumptions, which should be articulated.
        \item The authors should reflect on the factors that influence the performance of the approach. For example, a facial recognition algorithm may perform poorly when image resolution is low or images are taken in low lighting. Or a speech-to-text system might not be used reliably to provide closed captions for online lectures because it fails to handle technical jargon.
        \item The authors should discuss the computational efficiency of the proposed algorithms and how they scale with dataset size.
        \item If applicable, the authors should discuss possible limitations of their approach to address problems of privacy and fairness.
        \item While the authors might fear that complete honesty about limitations might be used by reviewers as grounds for rejection, a worse outcome might be that reviewers discover limitations that aren't acknowledged in the paper. The authors should use their best judgment and recognize that individual actions in favor of transparency play an important role in developing norms that preserve the integrity of the community. Reviewers will be specifically instructed to not penalize honesty concerning limitations.
    \end{itemize}

\item {\bf Theory Assumptions and Proofs}
    \item[] Question: For each theoretical result, does the paper provide the full set of assumptions and a complete (and correct) proof?
    \item[] Answer:  \answerNA{} %
    \item[] Justification: We have no theoretical results.
    \item[] Guidelines:
    \begin{itemize}
        \item The answer NA means that the paper does not include theoretical results. 
        \item All the theorems, formulas, and proofs in the paper should be numbered and cross-referenced.
        \item All assumptions should be clearly stated or referenced in the statement of any theorems.
        \item The proofs can either appear in the main paper or the supplemental material, but if they appear in the supplemental material, the authors are encouraged to provide a short proof sketch to provide intuition. 
        \item Inversely, any informal proof provided in the core of the paper should be complemented by formal proofs provided in appendix or supplemental material.
        \item Theorems and Lemmas that the proof relies upon should be properly referenced. 
    \end{itemize}

    \item {\bf Experimental Result Reproducibility}
    \item[] Question: Does the paper fully disclose all the information needed to reproduce the main experimental results of the paper to the extent that it affects the main claims and/or conclusions of the paper (regardless of whether the code and data are provided or not)?
    \item[] Answer: \answerYes{} %
    \item[] Justification: We publicly release our code and data.
    \item[] Guidelines:
    \begin{itemize}
        \item The answer NA means that the paper does not include experiments.
        \item If the paper includes experiments, a No answer to this question will not be perceived well by the reviewers: Making the paper reproducible is important, regardless of whether the code and data are provided or not.
        \item If the contribution is a dataset and/or model, the authors should describe the steps taken to make their results reproducible or verifiable. 
        \item Depending on the contribution, reproducibility can be accomplished in various ways. For example, if the contribution is a novel architecture, describing the architecture fully might suffice, or if the contribution is a specific model and empirical evaluation, it may be necessary to either make it possible for others to replicate the model with the same dataset, or provide access to the model. In general. releasing code and data is often one good way to accomplish this, but reproducibility can also be provided via detailed instructions for how to replicate the results, access to a hosted model (e.g., in the case of a large language model), releasing of a model checkpoint, or other means that are appropriate to the research performed.
        \item While NeurIPS does not require releasing code, the conference does require all submissions to provide some reasonable avenue for reproducibility, which may depend on the nature of the contribution. For example
        \begin{enumerate}
            \item If the contribution is primarily a new algorithm, the paper should make it clear how to reproduce that algorithm.
            \item If the contribution is primarily a new model architecture, the paper should describe the architecture clearly and fully.
            \item If the contribution is a new model (e.g., a large language model), then there should either be a way to access this model for reproducing the results or a way to reproduce the model (e.g., with an open-source dataset or instructions for how to construct the dataset).
            \item We recognize that reproducibility may be tricky in some cases, in which case authors are welcome to describe the particular way they provide for reproducibility. In the case of closed-source models, it may be that access to the model is limited in some way (e.g., to registered users), but it should be possible for other researchers to have some path to reproducing or verifying the results.
        \end{enumerate}
    \end{itemize}

\item {\bf Open access to data and code}
    \item[] Question: Does the paper provide open access to the data and code, with sufficient instructions to faithfully reproduce the main experimental results, as described in supplemental material?
    \item[] Answer: \answerYes{} %
    \item[] Justification: We publicly release our code and data. The access instructions are in the supplementary material. The source Kaggle datasets can be accessed publicly. 
    \item[] Guidelines:
    \begin{itemize}
        \item The answer NA means that paper does not include experiments requiring code.
        \item Please see the NeurIPS code and data submission guidelines (\url{https://nips.cc/public/guides/CodeSubmissionPolicy}) for more details.
        \item While we encourage the release of code and data, we understand that this might not be possible, so “No” is an acceptable answer. Papers cannot be rejected simply for not including code, unless this is central to the contribution (e.g., for a new open-source benchmark).
        \item The instructions should contain the exact command and environment needed to run to reproduce the results. See the NeurIPS code and data submission guidelines (\url{https://nips.cc/public/guides/CodeSubmissionPolicy}) for more details.
        \item The authors should provide instructions on data access and preparation, including how to access the raw data, preprocessed data, intermediate data, and generated data, etc.
        \item The authors should provide scripts to reproduce all experimental results for the new proposed method and baselines. If only a subset of experiments are reproducible, they should state which ones are omitted from the script and why.
        \item At submission time, to preserve anonymity, the authors should release anonymized versions (if applicable).
        \item Providing as much information as possible in supplemental material (appended to the paper) is recommended, but including URLs to data and code is permitted.
    \end{itemize}

\item {\bf Experimental Setting/Details}
    \item[] Question: Does the paper specify all the training and test details (e.g., data splits, hyperparameters, how they were chosen, type of optimizer, etc.) necessary to understand the results?
    \item[] Answer: \answerYes{} %
    \item[] Justification: Our experimental settings are discussed in the paper with great details. We also release our code as the supplemental material.
    \item[] Guidelines:
    \begin{itemize}
        \item The answer NA means that the paper does not include experiments.
        \item The experimental setting should be presented in the core of the paper to a level of detail that is necessary to appreciate the results and make sense of them.
        \item The full details can be provided either with the code, in appendix, or as supplemental material.
    \end{itemize}

\item {\bf Experiment Statistical Significance}
    \item[] Question: Does the paper report error bars suitably and correctly defined or other appropriate information about the statistical significance of the experiments?
    \item[] Answer: \answerYes{} %
    \item[] Justification: For experiments, p-values and bar plots are included in the paper and the appendix.
    \item[] Guidelines:
    \begin{itemize}
        \item The answer NA means that the paper does not include experiments.
        \item The authors should answer "Yes" if the results are accompanied by error bars, confidence intervals, or statistical significance tests, at least for the experiments that support the main claims of the paper.
        \item The factors of variability that the error bars are capturing should be clearly stated (for example, train/test split, initialization, random drawing of some parameter, or overall run with given experimental conditions).
        \item The method for calculating the error bars should be explained (closed form formula, call to a library function, bootstrap, etc.)
        \item The assumptions made should be given (e.g., Normally distributed errors).
        \item It should be clear whether the error bar is the standard deviation or the standard error of the mean.
        \item It is OK to report 1-sigma error bars, but one should state it. The authors should preferably report a 2-sigma error bar than state that they have a 96\% CI, if the hypothesis of Normality of errors is not verified.
        \item For asymmetric distributions, the authors should be careful not to show in tables or figures symmetric error bars that would yield results that are out of range (e.g. negative error rates).
        \item If error bars are reported in tables or plots, The authors should explain in the text how they were calculated and reference the corresponding figures or tables in the text.
    \end{itemize}

\item {\bf Experiments Compute Resources}
    \item[] Question: For each experiment, does the paper provide sufficient information on the computer resources (type of compute workers, memory, time of execution) needed to reproduce the experiments?
    \item[] Answer: \answerYes{} %
    \item[] Justification: Our LLM annotation methods aim to reduce the costs of dataset annotation, and the estimated costs are in the Appendix.
    \item[] Guidelines:
    \begin{itemize}
        \item The answer NA means that the paper does not include experiments.
        \item The paper should indicate the type of compute workers CPU or GPU, internal cluster, or cloud provider, including relevant memory and storage.
        \item The paper should provide the amount of compute required for each of the individual experimental runs as well as estimate the total compute. 
        \item The paper should disclose whether the full research project required more compute than the experiments reported in the paper (e.g., preliminary or failed experiments that didn't make it into the paper). 
    \end{itemize}
    
\item {\bf Code Of Ethics}
    \item[] Question: Does the research conducted in the paper conform, in every respect, with the NeurIPS Code of Ethics \url{https://neurips.cc/public/EthicsGuidelines}?
    \item[] Answer: \answerYes{} %
    \item[] Justification: We have reviewed the NeurIPS Code of Ethics and confirm that our research conforms to it.
    \item[] Guidelines:
    \begin{itemize}
        \item The answer NA means that the authors have not reviewed the NeurIPS Code of Ethics.
        \item If the authors answer No, they should explain the special circumstances that require a deviation from the Code of Ethics.
        \item The authors should make sure to preserve anonymity (e.g., if there is a special consideration due to laws or regulations in their jurisdiction).
    \end{itemize}

\item {\bf Broader Impacts}
    \item[] Question: Does the paper discuss both potential positive societal impacts and negative societal impacts of the work performed?
    \item[] Answer: \answerNA{} %
    \item[] Justification: Dataset annotation is a standard component in all machine learning projects, and improvements on the dataset annotation process should bring no risks of negative societal impacts. 
    \item[] Guidelines:
    \begin{itemize}
        \item The answer NA means that there is no societal impact of the work performed.
        \item If the authors answer NA or No, they should explain why their work has no societal impact or why the paper does not address societal impact.
        \item Examples of negative societal impacts include potential malicious or unintended uses (e.g., disinformation, generating fake profiles, surveillance), fairness considerations (e.g., deployment of technologies that could make decisions that unfairly impact specific groups), privacy considerations, and security considerations.
        \item The conference expects that many papers will be foundational research and not tied to particular applications, let alone deployments. However, if there is a direct path to any negative applications, the authors should point it out. For example, it is legitimate to point out that an improvement in the quality of generative models could be used to generate deepfakes for disinformation. On the other hand, it is not needed to point out that a generic algorithm for optimizing neural networks could enable people to train models that generate Deepfakes faster.
        \item The authors should consider possible harms that could arise when the technology is being used as intended and functioning correctly, harms that could arise when the technology is being used as intended but gives incorrect results, and harms following from (intentional or unintentional) misuse of the technology.
        \item If there are negative societal impacts, the authors could also discuss possible mitigation strategies (e.g., gated release of models, providing defenses in addition to attacks, mechanisms for monitoring misuse, mechanisms to monitor how a system learns from feedback over time, improving the efficiency and accessibility of ML).
    \end{itemize}
    
\item {\bf Safeguards}
    \item[] Question: Does the paper describe safeguards that have been put in place for responsible release of data or models that have a high risk for misuse (e.g., pretrained language models, image generators, or scraped datasets)?
    \item[] Answer: \answerNA{} %
    \item[] Justification: Dataset annotation is a standard component in all machine learning projects and our work poses no risks of misuse.
    \item[] Guidelines:
    \begin{itemize}
        \item The answer NA means that the paper poses no such risks.
        \item Released models that have a high risk for misuse or dual-use should be released with necessary safeguards to allow for controlled use of the model, for example by requiring that users adhere to usage guidelines or restrictions to access the model or implementing safety filters. 
        \item Datasets that have been scraped from the Internet could pose safety risks. The authors should describe how they avoided releasing unsafe images.
        \item We recognize that providing effective safeguards is challenging, and many papers do not require this, but we encourage authors to take this into account and make a best faith effort.
    \end{itemize}

\item {\bf Licenses for existing assets}
    \item[] Question: Are the creators or original owners of assets (e.g., code, data, models), used in the paper, properly credited and are the license and terms of use explicitly mentioned and properly respected?
    \item[] Answer:  \answerYes{} %
    \item[] Justification: The assets we created are released with public licenses. The original licenses of the Kaggle datasets can be found on the dataset webpages, and we do not re-release their data or modify their licenses to conform to Kaggle's terms of use. Users may download the Kaggle datasets on their own through public Kaggle API access, following our process or using our tools. Kaggle datasets have licenses that permit data analysis usages.
    \item[] Guidelines:
    \begin{itemize}
        \item The answer NA means that the paper does not use existing assets.
        \item The authors should cite the original paper that produced the code package or dataset.
        \item The authors should state which version of the asset is used and, if possible, include a URL.
        \item The name of the license (e.g., CC-BY 4.0) should be included for each asset.
        \item For scraped data from a particular source (e.g., website), the copyright and terms of service of that source should be provided.
        \item If assets are released, the license, copyright information, and terms of use in the package should be provided. For popular datasets, \url{paperswithcode.com/datasets} has curated licenses for some datasets. Their licensing guide can help determine the license of a dataset.
        \item For existing datasets that are re-packaged, both the original license and the license of the derived asset (if it has changed) should be provided.
        \item If this information is not available online, the authors are encouraged to reach out to the asset's creators.
    \end{itemize}

\item {\bf New Assets}
    \item[] Question: Are new assets introduced in the paper well documented and is the documentation provided alongside the assets?
    \item[] Answer: \answerYes{} %
    \item[] Justification: Our model and dataset are well documented and released publicly.
    \item[] Guidelines:
    \begin{itemize}
        \item The answer NA means that the paper does not release new assets.
        \item Researchers should communicate the details of the dataset/code/model as part of their submissions via structured templates. This includes details about training, license, limitations, etc. 
        \item The paper should discuss whether and how consent was obtained from people whose asset is used.
        \item At submission time, remember to anonymize your assets (if applicable). You can either create an anonymized URL or include an anonymized zip file.
    \end{itemize}

\item {\bf Crowdsourcing and Research with Human Subjects}
    \item[] Question: For crowdsourcing experiments and research with human subjects, does the paper include the full text of instructions given to participants and screenshots, if applicable, as well as details about compensation (if any)? 
    \item[] Answer: \answerNA{} %
    \item[] Justification: The paper does not involve crowdsourcing nor research with human subjects.
    \item[] Guidelines:
    \begin{itemize}
        \item The answer NA means that the paper does not involve crowdsourcing nor research with human subjects.
        \item Including this information in the supplemental material is fine, but if the main contribution of the paper involves human subjects, then as much detail as possible should be included in the main paper. 
        \item According to the NeurIPS Code of Ethics, workers involved in data collection, curation, or other labor should be paid at least the minimum wage in the country of the data collector. 
    \end{itemize}

\item {\bf Institutional Review Board (IRB) Approvals or Equivalent for Research with Human Subjects}
    \item[] Question: Does the paper describe potential risks incurred by study participants, whether such risks were disclosed to the subjects, and whether Institutional Review Board (IRB) approvals (or an equivalent approval/review based on the requirements of your country or institution) were obtained?
    \item[] Answer: \answerNA{} %
    \item[] Justification: The paper does not involve crowdsourcing nor research with human subjects.
    \item[] Guidelines:
    \begin{itemize}
        \item The answer NA means that the paper does not involve crowdsourcing nor research with human subjects.
        \item Depending on the country in which research is conducted, IRB approval (or equivalent) may be required for any human subjects research. If you obtained IRB approval, you should clearly state this in the paper. 
        \item We recognize that the procedures for this may vary significantly between institutions and locations, and we expect authors to adhere to the NeurIPS Code of Ethics and the guidelines for their institution. 
        \item For initial submissions, do not include any information that would break anonymity (if applicable), such as the institution conducting the review.
    \end{itemize}

\end{enumerate}

\end{document}